\documentclass[10pt,twocolumn,letterpaper]{article}

\usepackage{cvpr} %

\usepackage{epsfig}
\usepackage{graphicx}
\usepackage{amsmath}
\usepackage{amssymb}
\usepackage{booktabs}
\usepackage{lib}

\usepackage{stmaryrd} %
\usepackage{paralist} %
\usepackage[page]{appendix} %

\usepackage[symbol]{footmisc}

\usepackage[pagebackref,breaklinks,colorlinks,urlcolor=red]{hyperref}

\usepackage{xfrac}  

\usepackage[capitalize]{cleveref}
\crefname{section}{Sec.}{Secs.}
\Crefname{section}{Section}{Sections}
\Crefname{table}{Table}{Tables}
\crefname{table}{Tab.}{Tabs.}

\usepackage{color}
\definecolor{crimson}{rgb}{0.86, 0.08, 0.24}
\definecolor{gray}{rgb}{0.5,0.5,0.5}
\definecolor{green}{rgb}{0, 0.4, 0}
\definecolor{orange}{rgb}{1, 0.5, 0}
\definecolor{mahogany}{rgb}{0.75, 0.25, 0.0}
\definecolor{purple}{rgb}{0.6, 0, 0.6}
\definecolor{darkgreen}{rgb}{0, 0.4, 0}
\definecolor{frenchblue}{rgb}{0.0, 0.45, 0.73}
\definecolor{blue}{rgb}{0.0, 0.0, 0.65}
\definecolor{red}{rgb}{1,0,0}
\definecolor{yellow}{rgb}{1,1,0}
\definecolor{magenta}{rgb}{1,0,1}
\definecolor{pink}{rgb}{1,0.412,0.706}
\definecolor{newgreen}{rgb}{0, 0.6, 0.2}

\newlength\paramargin
\newlength\figmargin
\newlength\subfigmargin
\newlength\subsecmargin
\newlength\tabmargin
\newlength\eqmargin

\newlength\presecmargin
\newlength\secmargin

\newcommand{\red}[1]{{\color{red}{#1}}}
\def \submission {}
\ifx \submission \undefined

\else

\fi

\def\cvprPaperID{448} %
\def\confName{CVPR}
\def\confYear{2022}

\begin{document}

\title{AutoSDF: Shape Priors for 3D Completion, Reconstruction, and Generation}

\author{
Paritosh Mittal$^{*,1}$
\and
Yen-Chi Cheng$^{*,1}$
\and
Maneesh Singh$^{2}$
\and
Shubham Tulsiani$^{1}$
\and
$^{1}$Carnegie Mellon University
\and
$^{2}$Verisk Analytics
\and
\small \url{https://yccyenchicheng.github.io/AutoSDF/}
}

\twocolumn[{%
\maketitle
\renewcommand\twocolumn[1][]{#1}%
    \centering 
    \vspace{-2.0mm}
    \includegraphics[width=1.0\linewidth]{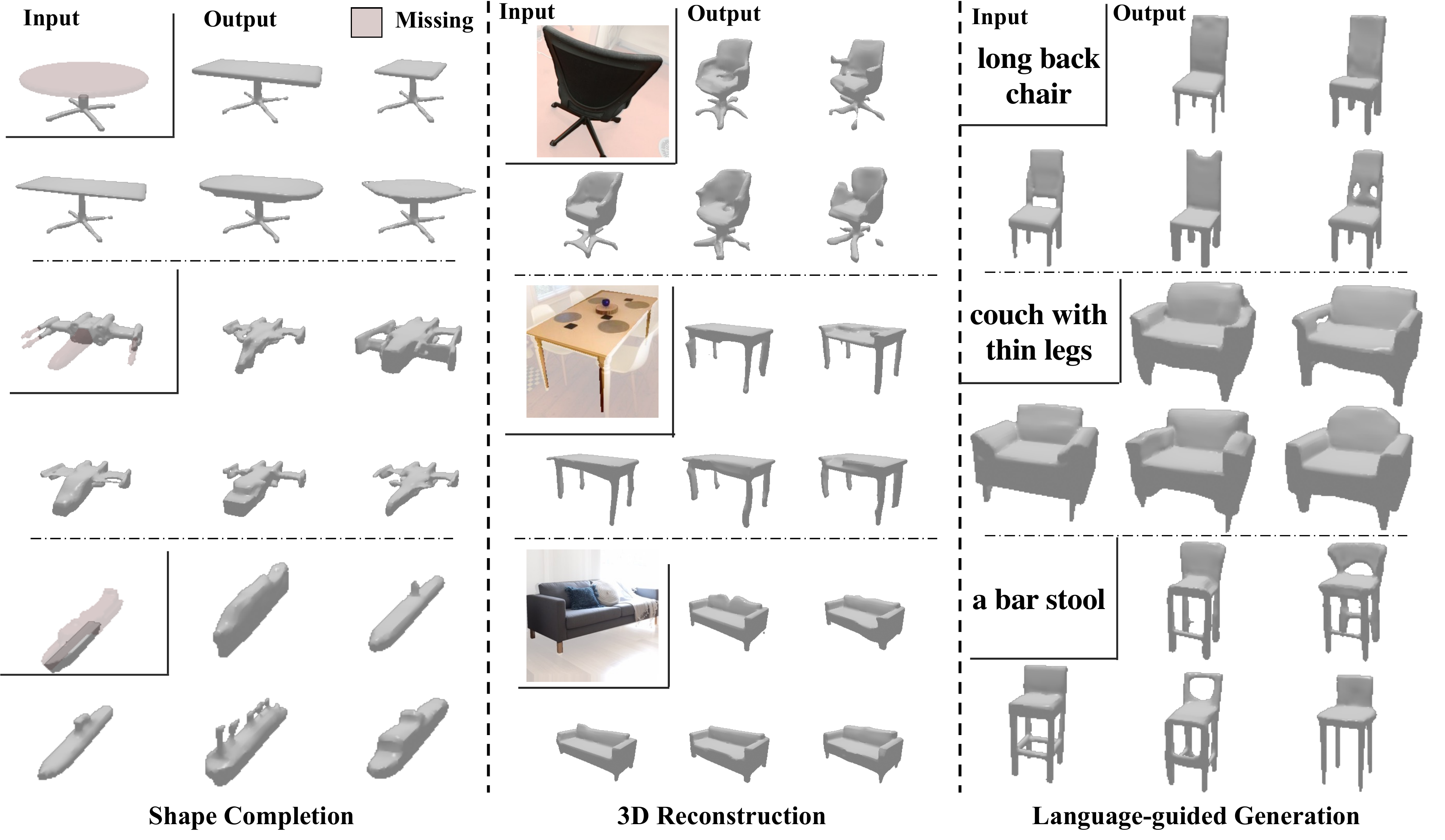}
    \captionof{figure}{
        Our approach combines a novel, non-sequential autoregressive prior, capturing the distribution over 3D shapes, with task-specific conditionals, to generate multiple plausible and high-quality shapes consistent with input conditioning. We show the efficacy of our approach across diverse tasks such as shape completion, single-view reconstruction and language-guided generation.
    } \figlabel{teaser}
    \vspace{3mm}
}]

\begin{abstract}
\footnotetext[1]{indicates equal contribution}
Powerful priors allow us to perform inference with insufficient information. In this paper, we propose an autoregressive prior for 3D shapes to solve multimodal 3D tasks such as shape completion, reconstruction, and generation. We model the distribution over 3D shapes as a non-sequential autoregressive distribution over a discretized, low-dimensional, symbolic grid-like latent representation of 3D shapes. 
This enables us to represent distributions over 3D shapes conditioned on information from an arbitrary set of spatially anchored query locations and thus perform shape completion in such arbitrary settings (\eg generating a complete chair given only a view of the back leg).  We also show that the learned autoregressive prior can be leveraged for conditional tasks such as single-view reconstruction and language-based generation. This is achieved by learning task-specific `naive' conditionals which can be approximated by light-weight models trained on minimal paired data. We validate the effectiveness of the proposed method using both quantitative and qualitative evaluation and show that the proposed method outperforms the specialized state-of-the-art methods trained for individual tasks.
\end{abstract}

\newcommand{\val}{\mathbf{v}}
\newcommand{\pos}{\mathbf{x}}
\newcommand{\posg}{g}
\newcommand{\R}{\mathbb{R}}

\section{Introduction}
\seclabel{intro}
\vspace{\secmargin}

3D representations are essential for applications in robotics, self-driving, virtual/augmented reality, and online marketplaces. This has led to an increasing number of diverse tasks that rely on effective 3D representations -- a robot might need to predict the shape of the objects it encounters, an artist may want to imagine what a `thin couch' would look like, or a woodworker may want to explore possible tabletop designs to match the legs they carved. A common practice for tackling these tasks, such as 3D completion or single-view prediction is to utilize task-specific data and train individual systems for each task, requiring a large amount of compute and data resources. 

While tasks such as shape completion or image-conditioned prediction are seemingly different, they require similar outputs -- a distribution over the plausible 3D structure conditioned on the corresponding input. A generalized notion of what `tables' are is useful for both predicting the full shape from the left half and imagining what a `tall round table' may look like. In this work, we operationalize this observation and show that a generic shape prior can be leveraged across these different inference tasks. In particular, we propose to learn an expressive autoregressive shape prior from abundantly available raw 3D data. This prior can then help augment the task-specific conditional distributions which require paired training data (\eg language-shape pairs), and significantly improve performance when such paired data is difficult to acquire.

Learning such a prior directly over the continuous and high-dimensional space of 3D shapes is computationally intractable. 
Inspired by recent approaches that overcome similar challenges for image synthesis, we first leverage discrete representation learning to compute discretized and low-dimensional representations for  3D shapes. This not only preserves the essential information for decoding high-quality outputs but also makes the training of autoregressive models tractable. Moreover, to learn such a prior for a broad set of tasks such as shape completion where arbitrary subsets maybe observed \eg 4 legs of a chair, we propose to learn a `non-sequential' autoregressive prior \ie one capable of using random subsets as conditioning. To enable this, we also enforce that the discrete elements over which this prior is learned are encoded independently. 

We then present a common framework for leveraging our learned prior for conditional generation tasks \eg single-view reconstruction or language-guided generation (see \figref{teaser}). Instead of modeling the complex conditional distribution directly, we propose to approximate it as a product of the prior and task-specific `naive' conditionals, the latter of which can be learned without extensive training data. Combined with the rich and expressive shape prior, we find that this unified and simple approach leads to improvements over task-specific state-of-the-art methods.

\begin{figure*}[t!]
    \centering
    \includegraphics[width=0.90\linewidth]{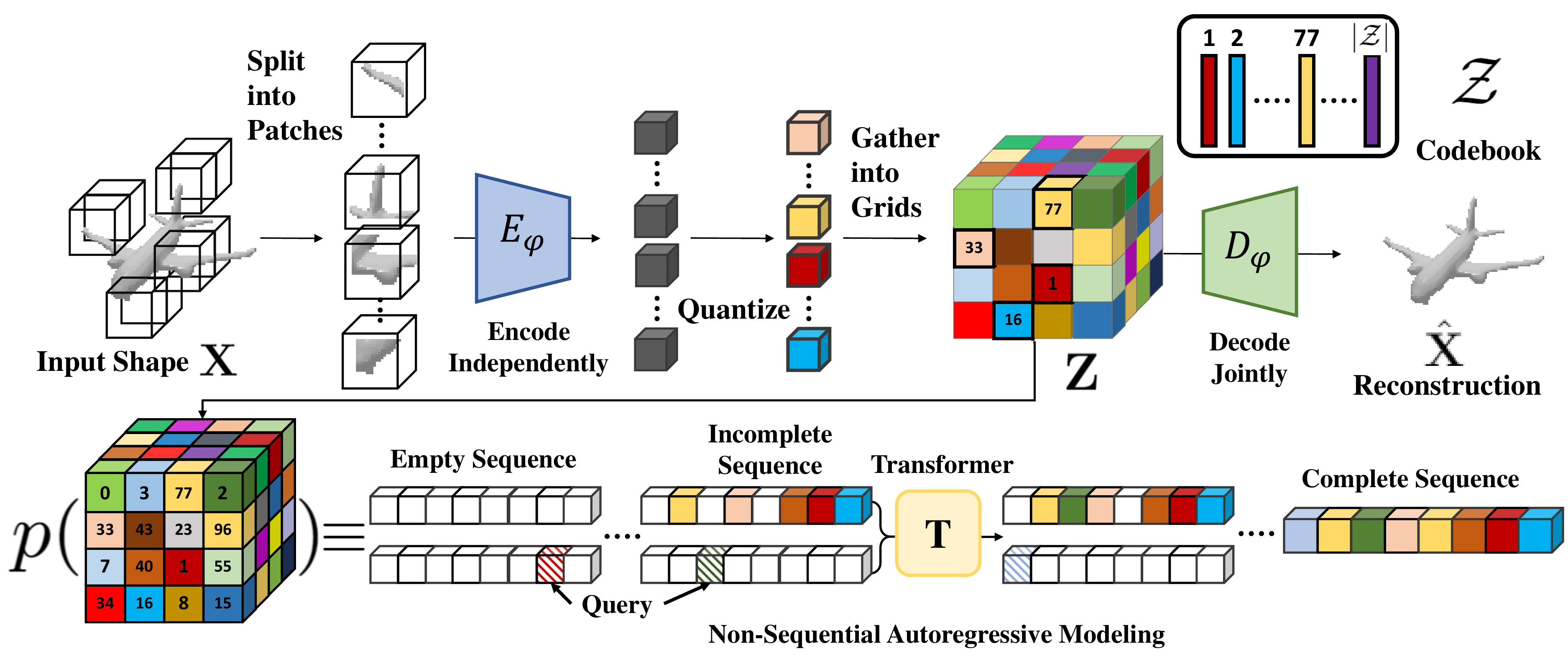}
    \caption{
    \textbf{Overview of Autoregressive Modeling.} \textbf{(top)} We use a VQ-VAE to extract a low-dimensional discrete representation of 3D shapes. Using a patch-wise encoder enables independently encoding local context and allows downstream tasks with partial observations. \textbf{(bottom)} We learn a transformer-based autoregressive model over the latent representation. Using randomized sampling orders allows learning a `non-sequential' autoregressive shape prior that can condition on arbitrary sets of partial latent observations.}
    \vspace \figmargin
    \figlabel{overview_p1}
\end{figure*} 
\section{Related Work}
\vspace{\secmargin}
\seclabel{related_work}
\paragraph{Autoregressive Modeling.}
Autoregressive models~\cite{van2016pixel,oord2016wavenet}  factorize the joint distribution over structured outputs into products of conditional distributions ($p(\mathbf{x}) = \Pi p(x_i|x_{<i}))$). Unlike GANs~\cite{goodfellow2014generative}, these can serve as powerful density estimators~\cite{NIPS2013_53adaf49}, are more stable during training ~\cite{NIPS2013_53adaf49, Salimans2017PixeCNN}, and can generalize well on held-out data. They have been successfully leveraged for modeling distributions across domains, such as images~\cite{van2016pixel, NIPS2016_b1301141,Salimans2017PixeCNN, chen2018pixelsnail}, audio~\cite{oord2016wavenet}, video~\cite{kalchbrenner2017video}, or language~\cite{yang2019xlnet}, and our work explores their benefits across a broad range of  3D generation tasks.

Following their recent successes in autoregressive modeling~\cite{yang2019xlnet,NEURIPS2020_1457c0d6,parmar2018image,chen2020generative}, our work adapts a Transformer-based~\cite{vaswani2017attention} architecture. While these approaches typically assume a sequential sampling order, closer to our work, Tulsiani and Gupta \cite{tulsiani2021pixel} extend these to allow non-sequential conditioning, which is important for tasks like completion. However, as they model distributions over low-level pixels, their approach cannot synthesize high-resolution outputs due to the quadratic complexity of Transformers. We therefore propose to first reduce high-dimensional 3D shapes to lower-dimensional discrete representations, and learn an autoregressive prior over this latent space.

We build on the work by van den Oord \etal~\cite{oord2017neural} who proposed a method to learn quantized and compact latent representations for images using Vector-Quantized Variational AutoEncoder (VQ-VAE), and later also introduced a hierarchical version~\cite{razavi2019generating}.
Inspired by Esser \etal~\cite {esser2021taming} who learned autoregressive generation over the discrete VQ-VAE representations, our work extends these ideas to the domain of 3D shapes. Different from these prior methods, we learn a non-sequential autoregressive prior while modifying the VQ-VAE architecture to independently encode the symbols, and show that this prior can be leveraged for downstream conditional inference tasks.

\vspace{\paramargin}
\paragraph{Shape Completion.} Completing full shapes from partial inputs such as discrete parts, or single-view 3D, is an increasingly important task across robotics and graphics. 
Most recent approaches~\cite{chen2019unpaired,zhang2021unsupervised,tchapmi2019topnet,yuan2018pcn,achlioptas2018learning,zhou20213d} formulate it as performing completion on point clouds and can infer plausible global shapes but have difficulty in either capturing fine-grained details, conditioning on sparse inputs, or generating diverse samples.
Our non-sequential autoregressive prior provides an alternative approach for shape completion. Given observations for an arbitrary (and possibly sparse) subregion of the 3D shape, we can sample diverse and high-quality shapes from our learned distribution, and we show that this generic approach performs comparably, if not better than previous specialized methods.

\vspace{\paramargin}
\paragraph{Single-view Reconstruction.} Inferring the 3D shape from a single image is an inherently ill-posed task -- an image of a chair from the back does not remove ambiguities about the shape of its seat. Several approaches have shown impressive single-view reconstruction results using voxels~\cite{drcTulsiani17,girdhar2016learning,wu2017marrnet,wu2018learning,choy20163d}, point clouds~~\cite{fan2017point,wu2020pq,mandikal20183d}, meshes~\cite{wang2018pixel2mesh,wang20193dn}, and most recently implicit  representations of 3D surfaces like  SDFs~\cite{jiang2020sdfdiff ,Xu2019disn}, UDFs~\cite{chibane2020udf} and CSPs~\cite{venkatesh2021csp}.
However, these are often deterministic in nature and only generate a 3D single output. By treating image-based prediction as conditional distributions that can be combined with a generic autoregressive prior, our work provides a simple and elegant way of inferring multiple plausible outputs, while also yielding empirical improvements. 
\vspace{\paramargin}
\paragraph{Language-based Generation.} Language is a highly effective and parsimonious modality for describing real-world shapes and objects. Chen \etal~\cite{chen2018text2shape} proposed a method to learn a joint text-shape embedding, followed by a GAN~\cite{goodfellow2014generative} based generator for synthesizing 3D from texts. However, generating shapes from texts is a fundamentally multi-modal task, and a GAN-based approach struggles to capture the multiple output modes. In contrast, learning naive-language guided conditional distributions from text aimed at disambiguation shapes~\cite{achlioptas2019shapeglot} and combining these with a generic prior, our work can generate diverse and plausible shapes.

\section{Approach}
\vspace{\secmargin}
\seclabel{approach}
We propose an autoregressive method to learn the distribution $p(\mathbf{X})$ over possible 3D shapes $\bfX$. Our method uses a volumetric Truncated-Signed Distance Field (T-SDF) for representing a 3D shape and learns a Transformer-based~\cite{vaswani2017attention} neural autoregressive model. However, as the computational complexity of transformers increases quadratically with the input dimension, we first map the high dimensional 3D shape to a corresponding low dimensional, discretized latent space. We then learn a `non-sequential' autoregressive prior over this compressed discrete representation, and show that this learned prior can be leveraged across diverse conditional generation tasks.

\begin{figure*}[t!]
    \centering
    \includegraphics[width=0.95\linewidth]{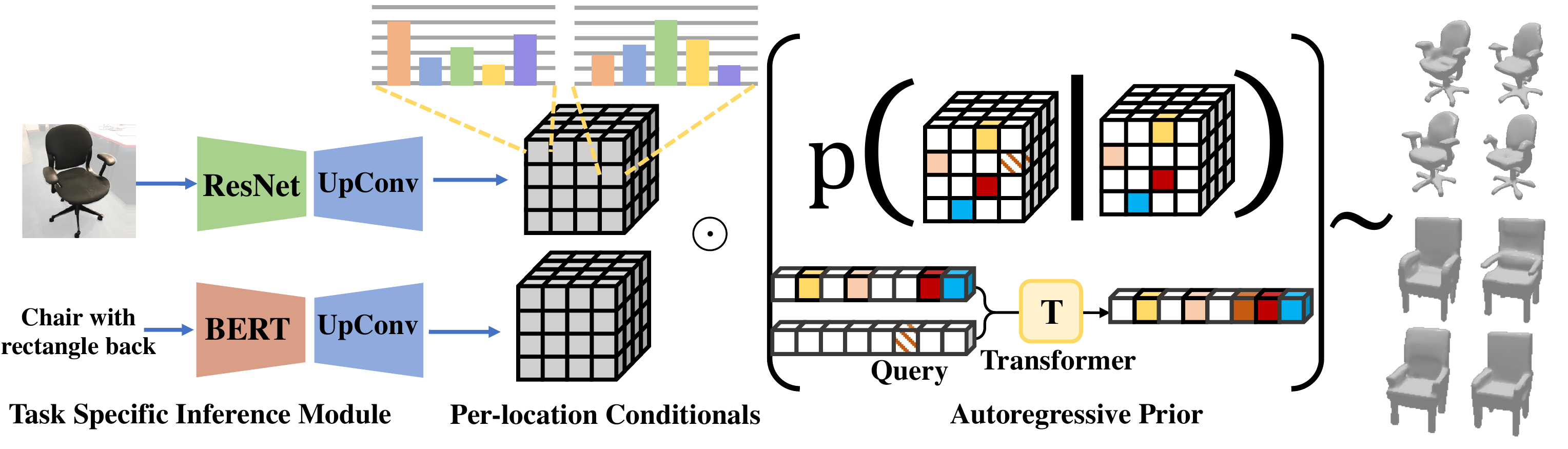}
    \vspace{-2mm}
    \caption{
    \textbf{Overview of conditional generation.} The proposed autoregressive prior can be used across diverse conditional generation tasks. For each task, we use a domain specific encoder followed by 3D up-convolutions to learn task specific conditional distributions. During inference, we can sample from the product distribution of the predicted conditionals and the learned autoregressive prior.    %
    }
    \vspace \figmargin
    \figlabel{overview_p2}
\end{figure*} 

\subsection{Discretized Latent Space for 3D Shapes}
\vspace{\subsecmargin}
\noindent To learn an effective autoregressive model, we aim to reduce the high-dimensional continuous 3D shape representation to a lower-dimensional discrete latent space. Towards this, we adapt the VQ-VAE~\cite{oord2017neural} framework and learn a 3D-VQ-VAE whose encoder $E_\psi$ can compute the desired low-dimensional representation, and the decoder $D_\psi$ can map this latent space back to 3D shapes. Given a 3D shape $\bf X$ with spatial dimension of $D^3$, we have
\begin{equation}
    \begin{aligned}
        \mathbf{Z} = VQ(E_\psi(\mathbf{X})),~~\mathbf{X}' = D_\psi(\mathbf{Z}), \\
        \text{where}\,\,\, \mathbf{Z} \in \{1,\cdots,|\mathcal{Z}|\}^{d^3}
    \end{aligned}
\end{equation}
where $VQ$ is the Vector Quantization step that maps a vector to the nearest element in the codebook $\mathcal{Z}$ which is jointly learned while training the VQ-VAE~\cite{oord2017neural}. The latent representation $\mathbf{Z}$ is thus a 3D grid of elements from the codebook, and can equivalently be thought of as a grid of indices referring to the corresponding codebook entry. We use $z_{\mathbf{i}}$ to denote the latent variable in the grid $\mathbf{Z}$ at position $\mathbf{i}$.

While the above framework allows learning a compact and quantized latent space, the encoder \emph{jointly} processes an input shape, and thus can use a large receptive field to encode each latent symbol. Unfortunately, this is not a desirable property for tasks such as shape completion since the latent codes for encoded partial shapes may differ significantly from those of the encoded full shape -- thus  partial observations of shape may not correspond to partial observations of latent variables. To overcome this challenge, we propose Patch-wise Encoding VQ-VAE or P-VQ-VAE that encodes the local shape regions \emph{independently}, while decoding them \emph{jointly} -- this allows the discrete encodings to only depend on local context, while still allowing the decoder to reason more globally when generating a 3D shape. We visualize this proposed architecture in \figref{overview_p1}, and train it using a combination of three losses proposed by van den Oord \etal~\cite{oord2017neural}: reconstruction loss, the vector quantization objective, and the commitment loss .

\subsection{Non-sequential Autoregressive Modeling}
\vspace{\subsecmargin}
\seclabel{nonseq-modeling}
The latent space $\mathbf{Z}$ is a 3D grid of tokens representing the original 3D shape.
We can thus reduce the task of learning the distribution over continuous 3D shapes to learning $p(\mathbf{Z})$, which is a distribution over the lower-dimensional discrete space. Assuming some ordering of the latent variables \eg a raster scan, typical autoregressive model can approximate this distribution by factorizing it as a product of location specific conditionals:
$p(\mathbf{Z}) = \prod_{\mathbf{i}=[1,1,1]}^{[d,d,d]} p(z_{\mathbf{i}} | z_{<\mathbf{i}})$.

However, this factorization assumes a fixed ordering in which the tokens are observed/generated. More specifically, this factorization implies that we need to know all $z_{<\mathbf{i}}$ before we predict the `next' symbol $z_\mathbf{i}$. However, such  conditioning is not always possible. For example, if we only observe the wheels of a car, the corresponding symbols would not be the first $k$ elements in a predefined sequence but rather occupy some spatially arbitrary locations. To allow for such arbitrary conditioning in our autoregressive model, we propose an autoregressive model which can predict a categorical distribution over tokens conditioned on a random input sequence, and use the term `non-sequential' autoregressive model to highlight this capability.

We follow the observation from~\cite{tulsiani2021pixel} that  the joint distribution $P(\mathbf{Z})$ can be factorized into terms of the form $p(z_{\mathbf{i}} | \mathbf{O})$, where $\mathbf{O}$ is a random set of observed variables. As illustrated in \figref{overview_p1}, 
instead of using a rasterized sampling order, we use a randomly permuted sequence of latent variables $\{z_{\mathbf{g}_1}, z_{\mathbf{g}_2}, z_{\mathbf{g}_3}, \cdots\}$ for autoregressively modeling the distribution $P(\mathbf{Z})$:
\begin{equation}
    \begin{aligned}
        p(\mathbf{Z}) = p_{\theta}(z_{\mathbf{g}_1} | \varnothing) \cdot p_{\theta}(z_{\mathbf{g}_2} | z_{\mathbf{g}_1}) \cdot p_{\theta}(z_{\mathbf{g}_3} | z_{\mathbf{g}_1},z_{\mathbf{g}_2}) \cdots
    \end{aligned}
\end{equation}
We model the distribution $p_{\theta}(z_{\mathbf{i}} | \mathbf{O})$ using a transformer-based architecture which is parameterized by $\theta$ and takes in an arbitrary set $\mathbf{O} \equiv (\{\mathbf{g}_j\}_{j=1}^{k})$ of observed latent variables with known locations and predicts the categorical distribution for an arbitrary query location $i$. We learn this model by simply maximizing the log-likelihood of the encoded latent representations using randomized orders for autoregressive generation. The non-sequential autoregressive network models the distribution over the latent variables $\mathbf{Z}$, which can be mapped to full 3D shapes $\hat{\bfX} = D_\psi(\mathbf{Z})$. Please see appendix for details.

\subsection{Conditional Generation}
\vspace{\subsecmargin}
\seclabel{shapecomp}
Given the autoregressive model trained to predict the distribution over the latent representation of 3D shapes, we can leverage it to solve various conditional prediction tasks like shape completion, or generation based on modalities like image and language.

\vspace{\paramargin}
\paragraph{Shape Completion.} The proposed P-VQ-VAE encodes local regions independently. This enables us to map partially observed shape $\mathbf{X_p}$ to corresponding observed latent variables $\mathbf{O} = \{z_{\mathbf{g}_1},z_{\mathbf{g}_2},\cdots, z_{\mathbf{g}_k}\}$. Although these observations can be at arbitrary spatial locations, our transformer-based autoregressive model is specifically trained to handle such inputs. In particular, we can formulate the task of shape completion as:
\begin{equation}
    \begin{aligned} \eqlabel{shape_comp}
        p(\mathbf{X} | \mathbf{X_p}) \approx p(\mathbf{Z}| \mathbf{O}) = \prod_{j > k} p_{\theta}(z_{\mathbf{g}_j} | z_{\mathbf{g}_{<j}},\mathbf{O})
    \end{aligned}
\end{equation}
Based on the above formulation, we can directly use our model from \secref{nonseq-modeling} to autoregressively sample  complete latent codes from partial observations. These can then be converted to 3D shapes via the P-VQ-VAE decoder.

\vspace{\paramargin}
\paragraph{Approximating generic conditional distributions.}
While shape completion could be reduced to conditional inference given a partially observed latent code, this reduction does not apply to other generation tasks. More generally, we are interested in inferring shape distributions $p(\mathbf{X}|C)$, where $C$ represents some conditioning \eg an image, or a text description. Approximating this as a distribution over the latent space, our goal is to learn models for $p(\mathbf{Z}|C)$. A possible approach would be to model the terms of the full joint distribution $p(\mathbf{Z}|C) = \prod_{i} p(z_{\mathbf{i}}| z_{<\mathbf{i}},C)$. However, in absence of abundant training data, learning this complex joint distribution may not be feasible.

Instead of modeling this complex distribution, we make a simplifying assumption and propose to model this joint distribution as a product of the shape prior, coupled with independent `naive' conditional terms that weakly capture the dependence on the conditioning $C$:
\begin{equation*}
    \begin{aligned} \eqlabel{cond_gen}
        \prod_j p(z_{\mathbf{g}_j} | z_{\mathbf{g}_{<j}},C) \approx \prod_j p_{\theta}(z_{\mathbf{g}_j} | z_{\mathbf{g}_{<j}}) \cdot \prod_j p_{\phi}(z_{\mathbf{g}_j} | C)
    \end{aligned}
\end{equation*}
This factorization corresponds to assuming a factor graph where the conditioning $C$ is connected to each latent variable $z_{\mathbf{i}}$ with only a pairwise potential $p(\mathbf{z}_{\mathbf{i}}|C)$. While this is an approximation of the more general case, it enables efficient learning and inference.

\vspace{\paramargin}
\paragraph{Learning Naive Conditionals.}
This per-location distribution intuitively corresponds to an independent `naive` conditional distribution for each variable in the latent representation. E.g., if the language described a `thin chair', this term may capture that we expect thin structures around legs. We model this distribution using a neural network parameterized by $\phi$, and can train this network using task-specific paired supervision. In particular, given $(\mathbf{X},C)$ pairs, we learn $\phi$ by maximizing the log-likelihood $\log p_{\phi} (z_{\mathbf{i}} | C)$ for each variable in the encoded shape $\mathbf{Z}$. As illustrated in \figref{overview_p2}, our task-specific network ($\phi$) comprises of domain-specific Encoders (\eg ResNet~\cite{he2016deep} for images; BERT~\cite{devlin2018bert} for language etc.) followed by up-convolutional decoders to the predict the conditional distribution over elements in $\mathbf{Z}$ \ie $p_{\phi}(z_\mathbf{i} | C)$. 

\vspace{\paramargin}
\paragraph{Prior-guided conditional inference.}
\seclabel{condgen}
Using the learned task-specific network to model the naive distributions $p_{\phi}(z_\mathbf{i}|C)$, we can use \eqlabel{cond_gen} to combine it with our autoregressive prior to obtain a conditional distribution over shapes which can be used for multimodal generation.

\begin{figure*}[h!]
    \centering
    \includegraphics[width=0.95\linewidth]{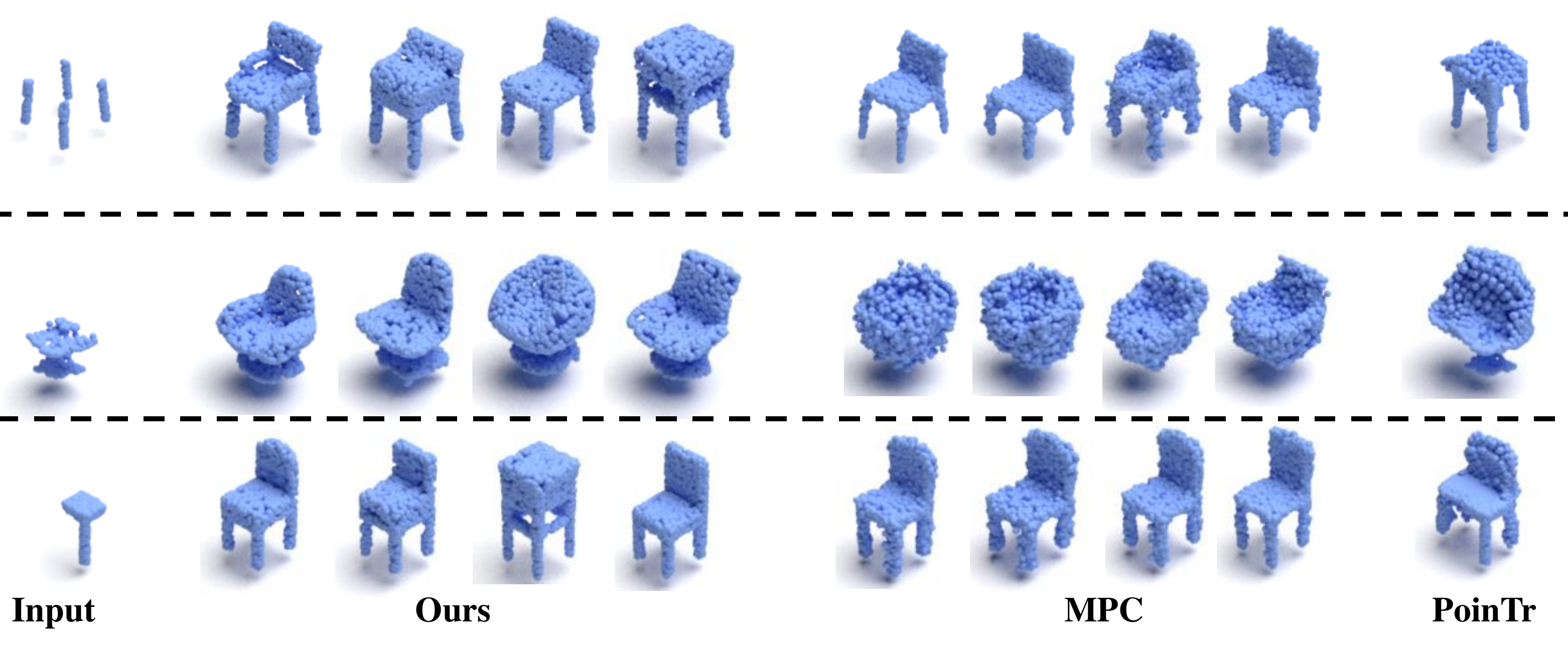}
    \caption{
    \textbf{Comparative results for Shape Completion.} Given the partial inputs, we visualize the generated results from different methods. Our approach yields more diverse  generations, while also better preserving the originally observed structure. For example, in the first row, given 4 slanted legs of a chair, some MPC generations make them straighter in the full point cloud, while they are preserved in our approach
    }
    \vspace \figmargin
    \figlabel{qual_shape_comp_base}
\end{figure*} 
\begin{figure*}[h!]
    \centering
    \includegraphics[width=0.95\linewidth]{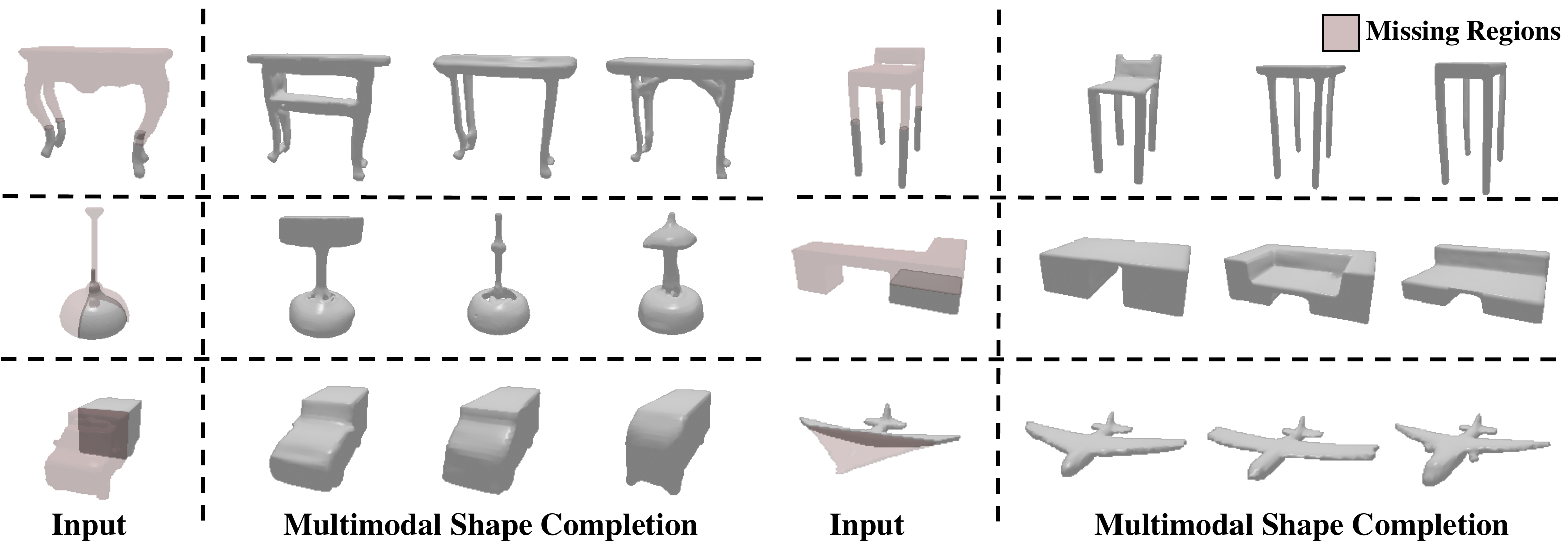}
    \caption{
    \textbf{Qualitative results for Shape Completion.} Our proposed approach is able to generate diverse plausible 3D shapes consistent with the partial input. The generated shapes are visually consistent with realistic shapes even with significantly missing parts( in \textbf{\red{Red}})}
    \vspace \figmargin
    \figlabel{qual_shape_comp}
\end{figure*} 
\begin{figure*}[t!]
    \centering
    \includegraphics[width=\linewidth]{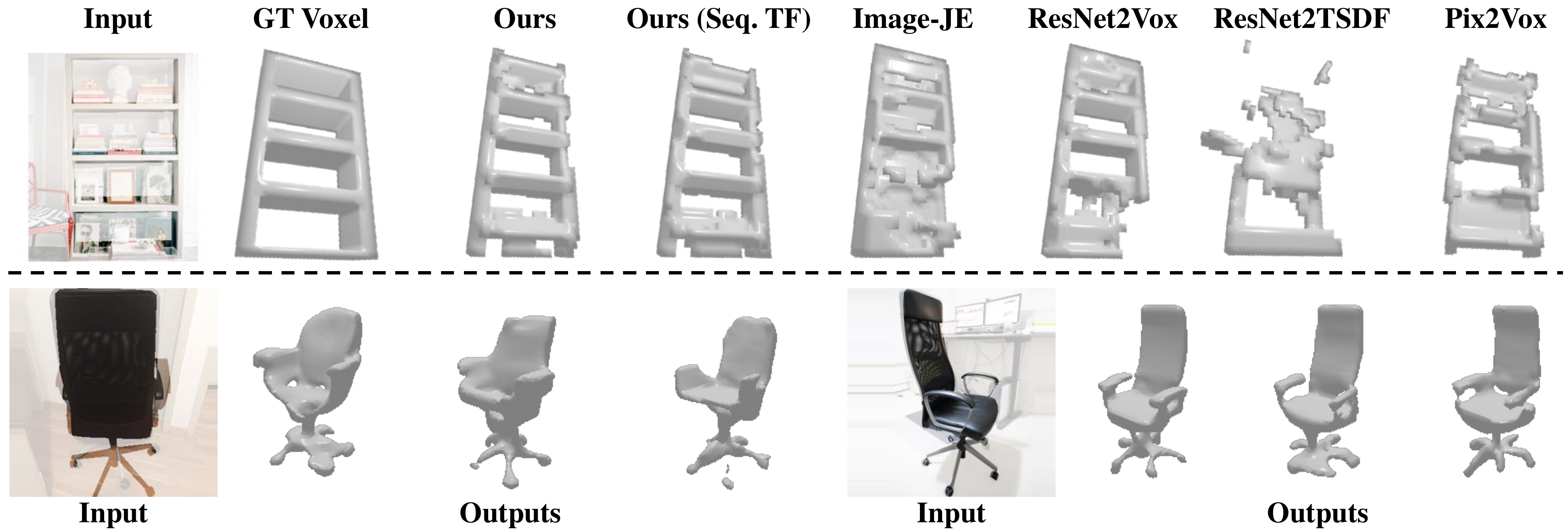}
    \caption{
    \textbf{Single-view 3D reconstruction}. (\textbf{Top}) We show a sample single-view reconstruction using our approach and other baselines -- our shape prior helps the generated shape be more globally coherent. (\textbf{Bottom}) We visualize multiple shapes predicted by our approach given the input images. We observe meaningful shape variation in the unobserved regions \eg front of the chair in the left image.
    }
    \vspace \figmargin
    \figlabel{qual_single_view}
\end{figure*} 
\begin{table}[t]
\caption{
    \textbf{Quantitative comparison on Shape Completion}.
}
\vspace{\tabmargin}
\centering
\setlength{\tabcolsep}{0.25em}
\small
\begin{tabular}{l cc cc cc} 
    \toprule
    & \multicolumn{2}{c}{Bottom Half} & \multicolumn{2}{c}{Octant} \\
    \cmidrule(r){2-3} \cmidrule(r){4-5}
    Method & \small{UHD $\shortdownarrow$} & \small{TMD $\shortuparrow$} & \small{UHD $\shortdownarrow$} & \small{TMD $\shortuparrow$} &  \\
    \midrule
    MPC~\cite{wu2020multimodal} & 0.0627 & 0.0303 & 0.0579 & 0.0376 \\
    PoinTr~\cite{yu2021pointr} & 0.0572 & N/A & \textbf{0.0536} & N/A \\
    Ours  & \textbf{0.0567} & \textbf{0.0341} & 0.0599 &\textbf{ 0.0693} \\
    \bottomrule
\end{tabular}
\vspace{\tabmargin}
\tablelabel{quan_shape_comp_less}
\end{table}

\section{Experiments}
\vspace{\secmargin}
\seclabel{experiment}
To demonstrate the efficacy of our autoregressive prior for generic 3D tasks, we quantitatively and qualitatively evaluated our method on three tasks -- a) shape completion, b) 3D reconstruction, and, c) language-guided generation.

\subsection{Multi-modal Shape Completion}
\vspace{\subsecmargin}
Our learned non-sequential autoregressive prior can naturally be adapted to the task of shape completion. We encode partial observations (in the form of local TSDFs) to obtain discrete symbols via the patchwise VQ-VAE encoder for the seen regions and sample full shapes conditioned on these observed symbols using the autoregressive prior.

\vspace{\paramargin}
\paragraph{Baselines and Evaluation Setup.} We evaluate our approach on the ShapeNet~\cite{shapenet2015} dataset using the train/test splits provided by Xu~\etal~\cite{Xu2019disn}. We use two completion settings for evaluation with varying fraction of observed shapes:
\begin{compactitem}
    \item \textit{Bottom half} of complete shape as input
    \item \textit{Octant} with front, left and bottom half of complete shape as input
\end{compactitem}

We compare our generations against two state-of-the-art point-cloud completion methods, MPC~\cite{wu2020multimodal} and PoinTr~\cite{yu2021pointr}. The former can generate multiple plausible shapes given the partial input, whereas the latter generates only a single (more accurate) completion. Both the publicly available PointTr~\cite{yu2021pointr} and our approach can use a single model to handle genetic shape completion scenarios. However, MPC needs to train a \emph{separate} model to handle completion in each different scenario. As our approach uses a partial TSDF input, it can potentially `see' information (up to a small threshold) beyond the boundaries. For a fair comparison, we also give the baseline methods additional points within the truncation threshold.

\vspace{\paramargin}
\paragraph{Evaluation Metrics.}
We adopt the metrics from MPC~\cite{wu2020multimodal} for the quantitative evaluation. These are given below. For each partial shape, we generate $k(= 10)$ complete shapes.
\begin{compactitem}
    \item \textit{Completion fidelity}: we compute the average of \textit{Unidirectional Hausdorff Distance} (UHD) from the input partial shape to the $k$ generated shapes. This measures the completion fidelity given the partial inputs.
    \item \textit{Completion diversity}: given $k$ generated results for each shape, we compute the average Chamfer distance to other $k-1$ shapes. The sum of the average distance among $k$ generation assesses the completion diversity and is denoted as \textit{Total Mutual Difference} (TMD).
\end{compactitem}
\vspace{\paramargin}
\paragraph{Results.}
To compare the performance of our approach with the  baselines on the task of shape completion, we use a set of held-out chairs from the ShapeNet dataset. Please note that while all the methods are trained across all ShapeNet classes, the evaluation is performed on a limited subset for computational reasons. The quantitative results reported  in~\tableref{quan_shape_comp_less} demonstrate that our autoregressive-prior based completion method performs favorably against the baselines, both in terms of fidelity and diversity on the two protocols.

We also show qualitative comparisons to these baselines in \figref{qual_shape_comp_base} and observe that our approach yields more diverse generations, while also better preserving the originally observed structure \eg given 4 slanted legs of a chair, some MPC generations make them straighter in the full point cloud, while the slants are preserved across the shapes generated by our approach. More shape completion results across other diverse shapes are shown in \figref{qual_shape_comp}. Our approach, while appropriately conditioning on the partial observations, generates a rich variety of diverse,  high-quality and realistic 3D shapes. It is noteworthy that although our autoregressive model is trained only on random observation sequences, it is able to condition on structured partial observations (anchored on a correlated set of anchored locations) that it was never trained explicitly to complete (unlike the point cloud completion baselines which are specifically trained for this task).

\subsection{Single-view 3D Prediction}
\vspace{\subsecmargin}
We next show that the learned prior can be leveraged for the task of single-view 3D reconstruction. To obtain the per-location image-conditioning, we train a modified ResNet~\cite{he2016deep} using pairs of images and corresponding encoded 3D models from the training dataset.

\begin{table}[h]
\caption{
    \textbf{Quantitative evaluation of single-view reconstruction}.
}
\vspace{\tabmargin}
\centering
\setlength{\tabcolsep}{0.25em}
\footnotesize
\begin{tabular}{l  ccc ccc} 
    \toprule
     & \multicolumn{3}{c}{ShapeNet} & \multicolumn{3}{c}{Pix3D} \\
    \cmidrule(r){2-4} \cmidrule(r){5-7}
    {Method} & {IoU $\shortuparrow$ } & {CD $\shortdownarrow$ }  & {F-Score $\shortuparrow$ } & {IoU $\shortuparrow$ } & {CD $\shortdownarrow$ }  & {F-Score $\shortuparrow$ }  \\
    \midrule
    Pix2Vox      & 0.467 & 2.521 & 0.335 & 0.504 & 3.001 & 0.385 \\
    ResNet2TSDF     & 0.478 & 2.684 & 0.320 & 0.475 & 4.582 & 0.351  \\
    ResNet2Voxel   & 0.457 & 2.501 & 0.316 & 0.505 & 4.670 & 0.357 \\
    Image-JE & 0.486 & 1.972 & 0.338 & 0.480 & 2.983 & 0.394 \\
    Ours (Sequential) & 0.554 & 1.448 & 0.393 & 0.516 & \textbf{2.254} & 0.412 \\
    Ours    &  \textbf{0.577} & \textbf{1.331} & \textbf{0.414} & \textbf{0.521} & 2.267 & \textbf{0.415} \\
    \bottomrule
\end{tabular}

\vspace{\tabmargin}
\tablelabel{reb_quan_svr}
\end{table}

\vspace{\paramargin}
\paragraph{Evaluation Setups.} We evaluate the proposed method on the ShapeNet rendered images~\cite{choy20163d}, and the real-world benchmark Pix3D~\cite{sun2018pix3d}, using cropped and segmented images as input for reconstruction. For ShapeNet, we use the same train/test split provided by Xu~\etal~\cite{Xu2019disn}, and evaluate against the voxelized models provided by Choy~\etal~\cite{choy20163d}. For Pix3D, we use the provided train/test splits for the chair category. In the absence of official splits for other categories, we randomly split the dataset into disjoint 3D shapes for training and testing. We evaluate all methods on the ground truth voxels in Pix3D and follow the official implementation to downsample all predictions into $32^3$ of voxels for evaluation. We use 3D IoU, Chamfer Distance (CD), and F-score@$1\%$~\cite{what3d_cvpr19} as the metrics to measure the performance across different methods.

\vspace{\paramargin}
\paragraph{Baselines} We compare with the following methods:
\begin{compactitem}
    \item Pix2Vox~\cite{xie2019pix2vox}: a state-of-the-art approach for 3D reconstruction.
    \item ResNet2TSDF / ResNet2Voxel: baselines which use a similar ResNet to ours but directly decode the output shape without using any shape prior.
    \item Image-JE: a transformer-based baseline to predict $P(S_i | S_{< i}, I)$ jointly, where $I$ is the input image.
    \item Ours (Sequential): a variant of the proposed method where the transformer 
\end{compactitem}

\vspace{\paramargin}
\paragraph{Results.} Quantitative results in \tableref{reb_quan_svr} demonstrate that our proposed method performs favorably across almost all  categories. Please refer to In \figref{qual_single_view}, we show some representative results with comparisons against the competing methods (additional results are in the supplementary). More crucially, as shown in the second row of \figref{qual_single_view}, unlike baselines, our approach can generate multiple plausible shapes given an input image. For example, given an image with a back-view of a chair , our model produces diverse reconstruction results with meaningful variation in unobserved regions, such as different armrests or cushions with varied shapes. %

\subsection{Language-guided Generation}
\vspace{\subsecmargin}
\begin{figure*}[t!]
    \centering
    \includegraphics[width=0.95\linewidth]{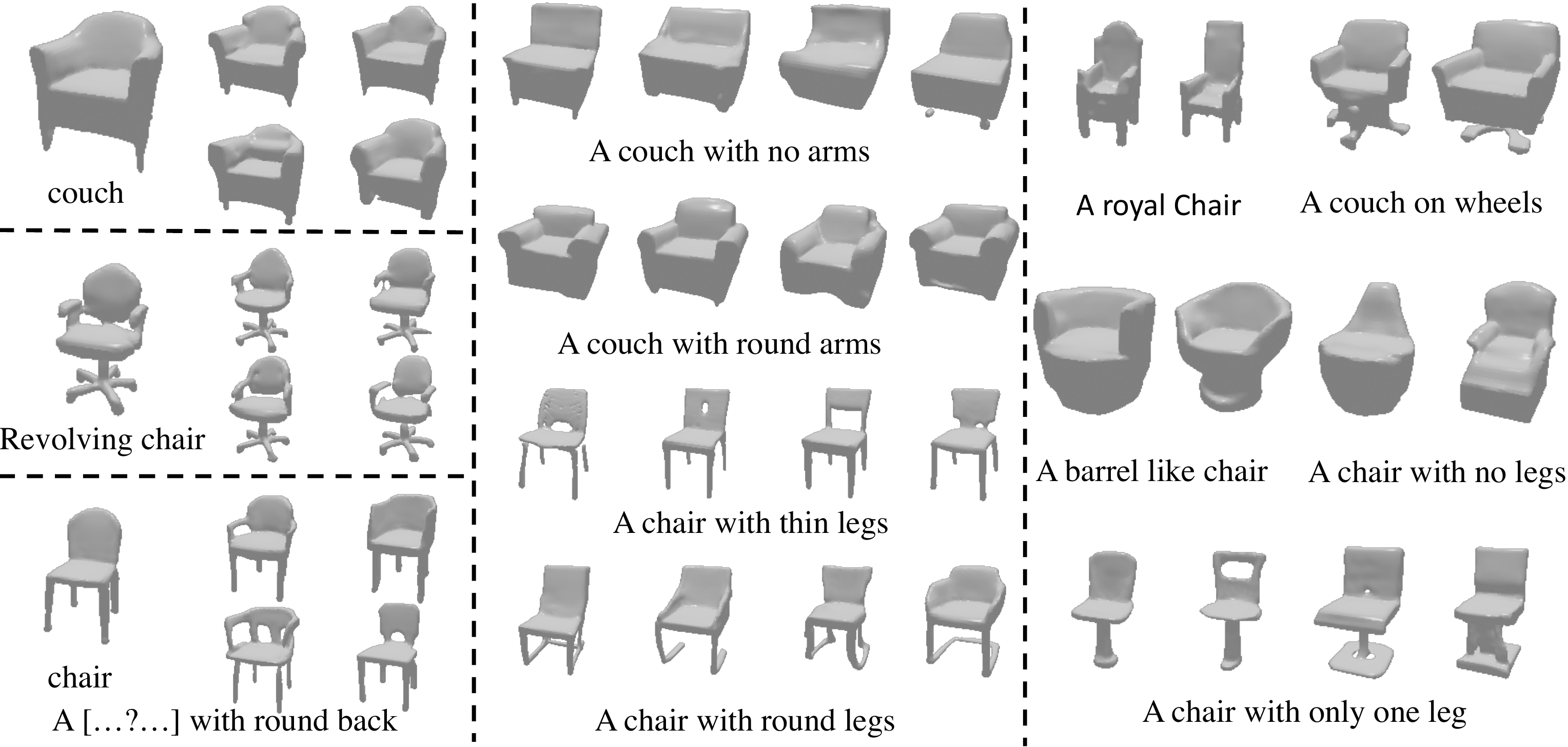}
    \caption{
    \textbf{Language Conditioned Generation}.
    The results signify that our approach can meaningfully estimate the correlation between input description and correspondingly plausible shapes while simultaneously generate the missing context required to generate them. 
    }
    \figlabel{quali_lang}
    \vspace \figmargin
\end{figure*}

Achlioptas \etal~\cite{achlioptas2019shapeglot} released a dataset containing text utterances describing the distinction between a target chair and the two distractors from Shapenet~\cite{shapenet2015}. We repurpose this data to train a text-conditioned generative model as described in \secref{condgen}. The distribution from this conditional when combined with our autoregressive prior, allows us to generate diverse shapes given a language description.

We compare our method with Text2Shape (T2S)~\cite{chen2018text2shape} on the text-guided 3D shape generation task. While T2S was originally trained to generate color and shape from text descriptions, after finetuning it on our dataset~\cite{achlioptas2019shapeglot}, we only use the generated shape for comparison.
In addition, we also compare our method with a transformer-based Encoder-Decoder Model (JE) trained on \cite{achlioptas2019shapeglot} to predict $P(S_i | S_{<i},T)$ jointly which serves as a baseline where a joint distribution is learned as opposed to our factored approach. Our approach combines two factors: a generic prior factor with an input modality dependent conditional factor. The latter may be potentially weak depending on the parsimony of the input modality and the amount of training data available. 

\vspace{\paramargin}
\paragraph{Quantitative Comparison.} To enable a comparison on this task, we train a neural evaluator similar to the one proposed in \cite{achlioptas2019shapeglot}. The evaluator is trained to distinguish the target shape from a distractor given  the specified text, and achieves a $\sim 83\%$ accuracy on this binary classification task. We use descriptions corresponding to held out test objects for evaluation. For each description, we provide the evaluator with two generations and report the quantitative results in \tableref{quan_lang}. We label an instance as `confused' when the absolute difference between the evaluator's confidence is $\leq 0.2$. We perform a seven way comparison by reporting the following comparisons -- Ours vs T2S, JE and Ours(Sequential), and, GT vs Ours, Ours (Sequential) , JE, and, T2S. We find that our approach is the preferred choice, with a very large margin, over either of the baselines (66:18 over T2S and 61:23 over JE). In a direct comparison with the GT, our generations are preferred 30\% of the time while the GT is preferred 49\%, and is significantly better than the other three baselines.

\vspace{\paramargin}
\paragraph{Qualitative Results.} While the quantitative evaluation above has been conducted  for a smaller set of  descriptions in the dataset, the model is trained over a larger set and can conditionally generate shapes for many more generic descriptions. We present an exemplary set of generated samples conditioned on a variety of input text in \figref{quali_lang}. The results clearly demonstrate that our approach can (a) generate highly plausible and realistic shapes correlated to the input description, (b) the shapes co-vary in a reasonable fashion with variations in the input text, \eg couch, revolving chair \etc, and, (c) that even when the descriptions refer only to a specific part \eg  `a chair with one leg', our model generates coherent global shapes consistent with this description. Please see appendix for additional results.

\begin{table}[t]
\caption{
    \textbf{Language-guided Generation}.
}
\vspace{\tabmargin}
\centering
\setlength{\tabcolsep}{0.25em}
\small
\begin{tabular}{cc cc cc} 
    \toprule
    Target~(Tr) & Distractor~(Dis) & $P(Tr)$  & $P(Dis)$ & $P(Conf)$ \\
    \midrule
    Ours & Text2Sshape~\cite{chen2018text2shape} & 66\% & 18\% & 16\% \\
    Ours & Joint-Encoding & 61\% & 23\% & 16\% \\
    Ours & Ours (Sequential) & 34\% & 27\% & 39\% \\
    Ours & GT & 30\% & 49\% & 21\% \\
    Ours (Sequential) & GT & 28\% & 52\% & 20\% \\
    Text2Sshape~\cite{chen2018text2shape} & GT & 15\% & 74\% & 11\% \\
    Joint-Encoding & GT & 19\% & 67\% & 14\% \\
    \bottomrule
\end{tabular}
\vspace{\tabmargin}
\tablelabel{quan_lang}
\end{table}
\vspace{-3mm}

\section{Discussion}
\seclabel{conclusion}
\vspace{ \secmargin }
We proposed an approach for learning a generic non-sequential autoregressive prior over 3D shapes useful  for \emph{multi-modal} generation for a diverse set of tasks \eg shape completion, single-view reconstruction, and language-guided synthesis. We find it encouraging that our unifying approach yields compelling results across these different tasks and is competitive with specifically designed baselines. However, a limitation of our conditional inference formulation is that it can only approximate the joint distribution -- while this helps in the low-paired data regime, this would be suboptimal with large-scale task-specific data. Moreover, our approach only applies to spatially structured 3D representations \eg TSDF or voxels, and it is not obvious whether  our autoregressive modeling framework can be adapted to other 3D representations such as meshes or neural implicit functions~\cite{shen2021deep}. Second, the proposed method might also be sensitive with respect to shape alignment. Finally, our learned shape prior is biased towards artificial categories with abundantly available CAD models, and cannot be leveraged for 3D generation beyond these.

{
\small
\bibliographystyle{ieee_fullname}
\bibliography{references}

\begin{thebibliography}{10}\itemsep=-1pt

\bibitem{achlioptas2018learning}
Panos Achlioptas, Olga Diamanti, Ioannis Mitliagkas, and Leonidas Guibas.
\newblock Learning representations and generative models for 3d point clouds.
\newblock In {\em ICML}, 2018.

\bibitem{achlioptas2019shapeglot}
Panos Achlioptas, Judy Fan, Robert Hawkins, Noah Goodman, and Leonidas~J
  Guibas.
\newblock Shapeglot: Learning language for shape differentiation.
\newblock In {\em CVPR}, 2019.

\bibitem{NEURIPS2020_1457c0d6}
Tom Brown, Benjamin Mann, Nick Ryder, Melanie Subbiah, Jared~D Kaplan, Prafulla
  Dhariwal, Arvind Neelakantan, Pranav Shyam, Girish Sastry, Amanda Askell,
  Sandhini Agarwal, Ariel Herbert-Voss, Gretchen Krueger, Tom Henighan, Rewon
  Child, Aditya Ramesh, Daniel Ziegler, Jeffrey Wu, Clemens Winter, Chris
  Hesse, Mark Chen, Eric Sigler, Mateusz Litwin, Scott Gray, Benjamin Chess,
  Jack Clark, Christopher Berner, Sam McCandlish, Alec Radford, Ilya Sutskever,
  and Dario Amodei.
\newblock Language models are few-shot learners.
\newblock In {\em NeurIPS}, 2020.

\bibitem{shapenet2015}
Angel~X. Chang, Thomas Funkhouser, Leonidas Guibas, Pat Hanrahan, Qixing Huang,
  Zimo Li, Silvio Savarese, Manolis Savva, Shuran Song, Hao Su, Jianxiong Xiao,
  Li Yi, and Fisher Yu.
\newblock {ShapeNet: An Information-Rich 3D Model Repository}.
\newblock Technical Report arXiv:1512.03012 [cs.GR], Stanford University ---
  Princeton University --- Toyota Technological Institute at Chicago, 2015.

\bibitem{chen2018text2shape}
Kevin Chen, Christopher~B Choy, Manolis Savva, Angel~X Chang, Thomas
  Funkhouser, and Silvio Savarese.
\newblock Text2shape: Generating shapes from natural language by learning joint
  embeddings.
\newblock In {\em ACCV}, 2018.

\bibitem{chen2020generative}
Mark Chen, Alec Radford, Rewon Child, Jeffrey Wu, Heewoo Jun, David Luan, and
  Ilya Sutskever.
\newblock Generative pretraining from pixels.
\newblock In {\em ICML}, 2020.

\bibitem{chen2019unpaired}
Xuelin Chen, Baoquan Chen, and Niloy~J Mitra.
\newblock Unpaired point cloud completion on real scans using adversarial
  training.
\newblock In {\em ICLR}, 2020.

\bibitem{chen2018pixelsnail}
Xi Chen, Nikhil Mishra, Mostafa Rohaninejad, and Pieter Abbeel.
\newblock Pixelsnail: An improved autoregressive generative model.
\newblock In {\em ICML}, 2018.

\bibitem{chibane2020udf}
Julian Chibane and Gerard Pons-Moll.
\newblock Neural unsigned distance fields for implicit function learning.
\newblock In {\em NeurIPS}, 2020.

\bibitem{choy20163d}
Christopher~B Choy, Danfei Xu, JunYoung Gwak, Kevin Chen, and Silvio Savarese.
\newblock 3d-r2n2: A unified approach for single and multi-view 3d object
  reconstruction.
\newblock In {\em ECCV}, 2016.

\bibitem{devlin2018bert}
Jacob Devlin, Ming-Wei Chang, Kenton Lee, and Kristina Toutanova.
\newblock Bert: Pre-training of deep bidirectional transformers for language
  understanding.
\newblock In {\em NAACL}, 2019.

\bibitem{esser2021taming}
Patrick Esser, Robin Rombach, and Bjorn Ommer.
\newblock Taming transformers for high-resolution image synthesis.
\newblock In {\em CVPR}, 2021.

\bibitem{fan2017point}
Haoqiang Fan, Hao Su, and Leonidas~J Guibas.
\newblock A point set generation network for 3d object reconstruction from a
  single image.
\newblock In {\em CVPR}, 2017.

\bibitem{girdhar2016learning}
Rohit Girdhar, David~F Fouhey, Mikel Rodriguez, and Abhinav Gupta.
\newblock Learning a predictable and generative vector representation for
  objects.
\newblock In {\em ECCV}, 2016.

\bibitem{goodfellow2014generative}
Ian Goodfellow, Jean Pouget-Abadie, Mehdi Mirza, Bing Xu, David Warde-Farley,
  Sherjil Ozair, Aaron Courville, and Yoshua Bengio.
\newblock Generative adversarial nets.
\newblock In {\em NeurIPS}, 2014.

\bibitem{he2016deep}
Kaiming He, Xiangyu Zhang, Shaoqing Ren, and Jian Sun.
\newblock Deep residual learning for image recognition.
\newblock In {\em CVPR}, 2016.

\bibitem{jiang2020sdfdiff}
Yue Jiang, Dantong Ji, Zhizhong Han, and Matthias Zwicker.
\newblock Sdfdiff: Differentiable rendering of signed distance fields for 3d
  shape optimization.
\newblock In {\em CVPR}, 2020.

\bibitem{kalchbrenner2017video}
Nal Kalchbrenner, A{\"a}ron Oord, Karen Simonyan, Ivo Danihelka, Oriol Vinyals,
  Alex Graves, and Koray Kavukcuoglu.
\newblock Video pixel networks.
\newblock In {\em ICML}, 2017.

\bibitem{mandikal20183d}
Priyanka Mandikal, Navaneet~K. L., Mayank Agarwal, and Venkatesh~Babu
  Radhakrishnan.
\newblock 3d-lmnet: Latent embedding matching for accurate and diverse 3d point
  cloud reconstruction from a single image.
\newblock In {\em BMVC}, 2018.

\bibitem{van2016pixel}
Aaron~Van Oord, Nal Kalchbrenner, and Koray Kavukcuoglu.
\newblock Pixel recurrent neural networks.
\newblock In {\em ICML}, 2016.

\bibitem{oord2016wavenet}
Aaron van~den Oord, Sander Dieleman, Heiga Zen, Karen Simonyan, Oriol Vinyals,
  Alex Graves, Nal Kalchbrenner, Andrew Senior, and Koray Kavukcuoglu.
\newblock Wavenet: A generative model for raw audio.
\newblock {\em arXiv preprint arXiv:1609.03499}, 2016.

\bibitem{oord2017neural}
Aaron van~den Oord, Oriol Vinyals, and Koray Kavukcuoglu.
\newblock Neural discrete representation learning.
\newblock In {\em NeurIPS}, 2017.

\bibitem{parmar2018image}
Niki Parmar, Ashish Vaswani, Jakob Uszkoreit, Lukasz Kaiser, Noam Shazeer,
  Alexander Ku, and Dustin Tran.
\newblock Image transformer.
\newblock In {\em ICML}, 2018.

\bibitem{razavi2019generating}
Ali Razavi, Aaron van~den Oord, and Oriol Vinyals.
\newblock Generating diverse high-fidelity images with vq-vae-2.
\newblock In {\em NeurIPS}, 2019.

\bibitem{Salimans2017PixeCNN}
Tim Salimans, Andrej Karpathy, Xi Chen, and Diederik~P. Kingma.
\newblock Pixelcnn++: A pixelcnn implementation with discretized logistic
  mixture likelihood and other modifications.
\newblock In {\em ICLR}, 2017.

\bibitem{shen2021deep}
Tianchang Shen, Jun Gao, Kangxue Yin, Ming-Yu Liu, and Sanja Fidler.
\newblock Deep marching tetrahedra: a hybrid representation for high-resolution
  3d shape synthesis.
\newblock {\em Advances in Neural Information Processing Systems}, 34, 2021.

\bibitem{sun2018pix3d}
Xingyuan Sun, Jiajun Wu, Xiuming Zhang, Zhoutong Zhang, Chengkai Zhang, Tianfan
  Xue, Joshua~B Tenenbaum, and William~T Freeman.
\newblock Pix3d: Dataset and methods for single-image 3d shape modeling.
\newblock In {\em CVPR}, 2018.

\bibitem{tancik2020fourier}
Matthew Tancik, Pratul~P Srinivasan, Ben Mildenhall, Sara Fridovich-Keil,
  Nithin Raghavan, Utkarsh Singhal, Ravi Ramamoorthi, Jonathan~T Barron, and
  Ren Ng.
\newblock Fourier features let networks learn high frequency functions in low
  dimensional domains.
\newblock In {\em NeurIPS}, 2020.

\bibitem{what3d_cvpr19}
Maxim Tatarchenko*, Stephan~R. Richter*, René Ranftl, Zhuwen Li, Vladlen
  Koltun, and Thomas Brox.
\newblock What do single-view 3d reconstruction networks learn?
\newblock 2019.

\bibitem{tchapmi2019topnet}
Lyne~P Tchapmi, Vineet Kosaraju, Hamid Rezatofighi, Ian Reid, and Silvio
  Savarese.
\newblock Topnet: Structural point cloud decoder.
\newblock In {\em CVPR}, 2019.

\bibitem{tulsiani2021pixel}
Shubham Tulsiani and Abhinav Gupta.
\newblock Pixeltransformer: Sample conditioned signal generation.
\newblock In {\em ICML}, 2021.

\bibitem{drcTulsiani17}
Shubham Tulsiani, Tinghui Zhou, Alexei~A. Efros, and Jitendra Malik.
\newblock Multi-view supervision for single-view reconstruction via
  differentiable ray consistency.
\newblock In {\em CVPR}, 2017.

\bibitem{NIPS2013_53adaf49}
Benigno Uria, Iain Murray, and Hugo Larochelle.
\newblock Rnade: The real-valued neural autoregressive density-estimator.
\newblock In {\em NeurIPS}, 2013.

\bibitem{NIPS2016_b1301141}
Aaron van~den Oord, Nal Kalchbrenner, Lasse Espeholt, koray kavukcuoglu, Oriol
  Vinyals, and Alex Graves.
\newblock Conditional image generation with pixelcnn decoders.
\newblock In {\em NeurIPS}, 2016.

\bibitem{vaswani2017attention}
Ashish Vaswani, Noam Shazeer, Niki Parmar, Jakob Uszkoreit, Llion Jones,
  Aidan~N Gomez, {\L}ukasz Kaiser, and Illia Polosukhin.
\newblock Attention is all you need.
\newblock In {\em NeurIPS}, 2017.

\bibitem{venkatesh2021csp}
Rahul Venkatesh, Tejan Karmali, Sarthak Sharma, Aurobrata Ghosh, R.~Venkatesh
  Babu, L\'aszl\'o~A. Jeni, and Maneesh Singh.
\newblock Deep implicit surface point prediction networks.
\newblock In {\em ICCV}, 2021.

\bibitem{wang2018pixel2mesh}
Nanyang Wang, Yinda Zhang, Zhuwen Li, Yanwei Fu, Wei Liu, and Yu-Gang Jiang.
\newblock Pixel2mesh: Generating 3d mesh models from single rgb images.
\newblock In {\em ECCV}, pages 52--67, 2018.

\bibitem{wang20193dn}
Weiyue Wang, Duygu Ceylan, Radomir Mech, and Ulrich Neumann.
\newblock 3dn: 3d deformation network.
\newblock In {\em CVPR}, 2019.

\bibitem{wu2017marrnet}
Jiajun Wu, Yifan Wang, Tianfan Xue, Xingyuan Sun, William~T Freeman, and
  Joshua~B Tenenbaum.
\newblock {MarrNet: 3D Shape Reconstruction via 2.5D Sketches}.
\newblock In {\em NeurIPS}, 2017.

\bibitem{wu2018learning}
Jiajun Wu, Chengkai Zhang, Xiuming Zhang, Zhoutong Zhang, William~T Freeman,
  and Joshua~B Tenenbaum.
\newblock Learning shape priors for single-view 3d completion and
  reconstruction.
\newblock In {\em ECCV}, 2018.

\bibitem{wu2020multimodal}
Rundi Wu, Xuelin Chen, Yixin Zhuang, and Baoquan Chen.
\newblock Multimodal shape completion via conditional generative adversarial
  networks.
\newblock In {\em ECCV}, 2020.

\bibitem{wu2020pq}
Rundi Wu, Yixin Zhuang, Kai Xu, Hao Zhang, and Baoquan Chen.
\newblock Pq-net: A generative part seq2seq network for 3d shapes.
\newblock In {\em CVPR}, 2020.

\bibitem{xie2019pix2vox}
Haozhe Xie, Hongxun Yao, Xiaoshuai Sun, Shangchen Zhou, and Shengping Zhang.
\newblock Pix2vox: Context-aware 3d reconstruction from single and multi-view
  images.
\newblock In {\em CVPR}, pages 2690--2698, 2019.

\bibitem{Xu2019disn}
Qiangeng Xu, Weiyue Wang, Duygu Ceylan, Radomir Mech, and Ulrich Neumann.
\newblock Disn: Deep implicit surface network for high-quality single-view 3d
  reconstruction.
\newblock In {\em NeurIPS}, 2019.

\bibitem{yang2019xlnet}
Zhilin Yang, Zihang Dai, Yiming Yang, Jaime Carbonell, Russ~R Salakhutdinov,
  and Quoc~V Le.
\newblock Xlnet: Generalized autoregressive pretraining for language
  understanding.
\newblock {\em NeurIPS}, 2019.

\bibitem{yu2021pointr}
Xumin Yu, Yongming Rao, Ziyi Wang, Zuyan Liu, Jiwen Lu, and Jie Zhou.
\newblock Poin{T}r: Diverse point cloud completion with geometry-aware
  transformers.
\newblock In {\em CVPR}, 2021.

\bibitem{yuan2018pcn}
Wentao Yuan, Tejas Khot, David Held, Christoph Mertz, and Martial Hebert.
\newblock Pcn: Point completion network.
\newblock In {\em 3DV}, 2018.

\bibitem{zhang2021unsupervised}
Junzhe Zhang, Xinyi Chen, Zhongang Cai, Liang Pan, Haiyu Zhao, Shuai Yi,
  Chai~Kiat Yeo, Bo Dai, and Chen~Change Loy.
\newblock Unsupervised 3d shape completion through gan inversion.
\newblock In {\em CVPR}, 2021.

\bibitem{zhou20213d}
Linqi Zhou, Yilun Du, and Jiajun Wu.
\newblock 3d shape generation and completion through point-voxel diffusion.
\newblock In {\em ICCV}, 2021.

\end{thebibliography}
}

\clearpage

\appendix

\twocolumn[{%
\renewcommand\twocolumn[1][]{#1}%
\centering \Large \textbf{Supplementary Materials: \\
AutoSDF: Shape Priors for 3D Completion, Reconstruction and Generation}
}]

\vspace{3mm}
We provide the implementation details of the P-VQ-VAE, transformer, image/text inference module, and the baselines such as ResNet2TSDF, ResNet2Voxel, Joint Text-Shape Baseline. For the visualization of the generated 3D shapes, please check our project page at \url{https://yccyenchicheng.github.io/AutoSDF/}.

\begin{table*}[t]
\caption{
    \textbf{Architecture for the P-VQ-VAE's encoder $E_{\psi}$}.
}
\vspace{-2mm}
\centering
\setlength{\tabcolsep}{1.0em}
\small
\begin{tabular}{c ccc ccc} 
    \toprule
    Layer name & Weights/Parameters & Input size & Output size \\
    \midrule
    Conv3D &  kernel $3 \times 3$, stride 1, padding 1 & $B \times 1 \times 8 \times 8 \times 8$ & $B \times 64 \times 8 \times 8 \times 8$ \\
    \midrule
    3D ResNet Block & kernel $3 \times 3$, stride 1, padding 1 & $B \times 64 \times 8 \times 8 \times 8$ & $B \times 64 \times 8 \times 8 \times 8$ \\
    \midrule
    Downsample & kernel $3 \times 3$, stride 2, padding 0 & $B \times 64 \times 8 \times 8 \times 8$ & $B \times 64 \times 4 \times 4 \times 4$ \\
    \midrule
    3D ResNet Block & kernel $3 \times 3$, stride 1, padding 1 & $B \times 64 \times 4 \times 4 \times 4$ & $B \times 128 \times 4 \times 4 \times 4$ \\
    \midrule
    Downsample & kernel $3 \times 3$, stride 2, padding 0 & $B \times 128 \times 4 \times 4 \times 4$ & $B \times 128 \times 2 \times 2 \times 2$ \\
    \midrule
    3D ResNet Block & kernel $3 \times 3$, stride 1, padding 1 & $B \times 128 \times 2 \times 2 \times 2$ & $B \times 128 \times 2 \times 2 \times 2$ \\
    \midrule
    Downsample & kernel $3 \times 3$, stride 2, padding 0 & $B \times 128 \times 2 \times 2 \times 2$ & $B \times 128 \times 1 \times 1 \times 1$ \\
    \midrule
    3D ResNet Block & kernel $3 \times 3$, stride 1, padding 1 & $B \times 128 \times 1 \times 1 \times 1$ & $B \times 256 \times 1 \times 1 \times 1$ \\
    \midrule
    3D ResNet Block & kernel $3 \times 3$, stride 1, padding 1 & $B \times 256 \times 1 \times 1 \times 1$ & $B \times 256 \times 1 \times 1 \times 1$ \\
    \midrule
    3D Attention Block & - & $B \times 256 \times 1 \times 1 \times 1$ & $B \times 256 \times 1 \times 1 \times 1$ \\
    \midrule
    3D ResNet Block & kernel $3 \times 3$, stride 1, padding 1 & $B \times 256 \times 1 \times 1 \times 1$ & $B \times 256 \times 1 \times 1 \times 1$ \\
    \midrule
    GroupNorm & num\_groups=32 & $B \times 256 \times 1 \times 1 \times 1$ & $B \times 256 \times 1 \times 1 \times 1$ \\
    \midrule
    Swish & - & $B \times 256 \times 1 \times 1 \times 1$ & $B \times 256 \times 1 \times 1 \times 1$ \\
    \midrule
    Conv3D & kernel $3 \times 3$, stride 1, padding 1 & $B \times 256 \times 1 \times 1 \times 1$ & $B \times 256 \times 1 \times 1 \times 1$ \\
    \bottomrule
\end{tabular}
\vspace{\tabmargin}
\tablelabel{tab_enc}
\end{table*}

\begin{table*}[t]
\caption{
    \textbf{Architecture for the P-VQ-VAE's decoder $D_{\psi}$}.
}
\vspace{-2mm}
\centering
\setlength{\tabcolsep}{1.0em}
\small
\begin{tabular}{c ccc ccc} 
    \toprule
    Layer name & Weights/Parameters & Input size & Output size \\
    \midrule
    Conv3D &  kernel $3 \times 3$, stride 1, padding 1 & $B \times 256 \times 1 \times 1 \times 1$ & $B \times 256 \times 1 \times 1 \times 1$ \\
    \midrule
    3D ResNet Block & kernel $3 \times 3$, stride 1, padding 1 & $B \times 256 \times 1 \times 1 \times 1$ & $B \times 256 \times 1 \times 1 \times 1$ \\
    \midrule
    3D Attention Block & - & $B \times 256 \times 1 \times 1 \times 1$ & $B \times 256 \times 1 \times 1 \times 1$ \\
    \midrule
    3D ResNet Block & kernel $3 \times 3$, stride 1, padding 1 & $B \times 256 \times 1 \times 1 \times 1$ & $B \times 256 \times 1 \times 1 \times 1$ \\
    \midrule
    Upsample & kernel $3 \times 3$, stride 2, padding 0 & $B \times 256 \times 1 \times 1 \times 1$ & $B \times 256 \times 2 \times 2 \times 2$ \\
    \midrule
    3D ResNet Block & kernel $3 \times 3$, stride 1, padding 1 & $B \times 256 \times 2 \times 2 \times 2$ & $B \times 128 \times 2 \times 2 \times 2$ \\
    \midrule
    3D Attention Block & - & $B \times 128 \times 2 \times 2 \times 2$ & $B \times 128 \times 2 \times 2 \times 2$ \\
    \midrule
    Upsample & kernel $3 \times 3$, stride 2, padding 0 & $B \times 128 \times 2 \times 2 \times 2$ & $B \times 128 \times 4 \times 4 \times 4$ \\
    \midrule
    3D ResNet Block & kernel $3 \times 3$, stride 1, padding 1 & $B \times 128 \times 4 \times 4 \times 4$ & $B \times 64 \times 4 \times 4 \times 4$ \\
    \midrule
    Upsample & kernel $3 \times 3$, stride 2, padding 0 & $B \times 64 \times 4 \times 4 \times 4$ & $B \times 64 \times 8 \times 8 \times 8$ \\
    \midrule
    3D ResNet Block & kernel $3 \times 3$, stride 1, padding 1 & $B \times 64 \times 8 \times 8 \times 8$ & $B \times 64 \times 8 \times 8 \times 8$ \\
    \midrule
    3D ResNet Block & kernel $3 \times 3$, stride 1, padding 1 & $B \times 64 \times 8 \times 8 \times 8$ & $B \times 64 \times 8 \times 8 \times 8$ \\
    \midrule
    GroupNorm & num\_groups=32 & $B \times 64 \times 8 \times 8 \times 8$ & $B \times 64 \times 8 \times 8 \times 8$ \\
    \midrule
    Swish & - & $B \times 64 \times 8 \times 8 \times 8$ & $B \times 64 \times 8 \times 8 \times 8$ \\
    \midrule
    Conv3D & kernel $3 \times 3$, stride 1, padding 1 & $B \times 64 \times 8 \times 8 \times 8$ & $B \times 1 \times 8 \times 8 \times 8$ \\
    \bottomrule
\end{tabular}
\vspace{\tabmargin}
\tablelabel{tab_dec}
\end{table*}

\begin{table*}[t]
\caption{
    \textbf{Architecture for the image inference module}.
}
\vspace{-2mm}
\centering
\setlength{\tabcolsep}{1.0em}
\small
\begin{tabular}{c ccc ccc} 
    \toprule
    Layer name & Weights/Parameters & Input size & Output size \\
    \midrule
    ResNet-18~\cite{he2016deep} &  - & $B \times 3 \times 256 \times 256$ & $B \times 512 \times 8 \times 8$ \\
    \midrule
    Linear\_to\_3D &  - & $B \times 512 \times 64$ & $B \times 512 \times 512$ \\
    \midrule
    3D ResNet Block & kernel $3 \times 3$, stride 1, padding 1 & $B \times 512 \times 8 \times 8 \times 8$ & $B \times 256 \times 8 \times 8 \times 8$ \\
    \midrule
    3D ResNet Block & kernel $3 \times 3$, stride 1, padding 1 & $B \times 256 \times 8 \times 8 \times 8$ & $B \times 128 \times 8 \times 8 \times 8$ \\
    \midrule
    3D ResNet Block & kernel $3 \times 3$, stride 1, padding 1 & $B \times 128 \times 8 \times 8 \times 8$ & $B \times 64 \times 8 \times 8 \times 8$ \\
    \midrule
    3D ResNet Block & kernel $3 \times 3$, stride 1, padding 1 & $B \times 64 \times 8 \times 8 \times 8$ & $B \times 64 \times 8 \times 8 \times 8$ \\
    \midrule
    Conv3D & kernel $3 \times 3$, stride 1, padding 1 & $B \times 64 \times 8 \times 8 \times 8$ & $B \times 512 \times 8 \times 8 \times 8$ \\
    \bottomrule
\end{tabular}
\vspace{\tabmargin}
\tablelabel{tab_img_arch}
\end{table*}

\begin{table*}[t]
\caption{
    \textbf{Architecture for the language inference module}.
}
\vspace{-2mm}
\centering
\setlength{\tabcolsep}{1.0em}
\small
\begin{tabular}{c ccc ccc} 
    \toprule
    Layer name & Weights/Parameters & Input size & Output size \\
    \midrule
    BERT~\cite{devlin2018bert} &  12 Layers, 12 Attention heads, 768 hidden dim & $B \times Seq$ & $B \times 768$ \\
    \midrule
    Linear\_up &  - & $B \times 768$ & $B \times 1024$ \\
    \midrule
    Linear\_down &  - & $B \times 1024$ & $B \times 512$ \\
    \midrule
    Conv3D &  kernel $3 \times 3$, stride 1, padding 1 & $B \times 1 \times 8 \times 8 \times 8$ & $B \times 64 \times 8 \times 8 \times 8$ \\
    \midrule
    3D ResNet Block & kernel $3 \times 3$, stride 1, padding 1 & $B \times 64 \times 8 \times 8 \times 8$ & $B \times 128 \times 8 \times 8 \times 8$ \\
    \midrule
    3D ResNet Block & kernel $3 \times 3$, stride 1, padding 1 & $B \times 128 \times 8 \times 8 \times 8$ & $B \times 256 \times 8 \times 8 \times 8$ \\
    \midrule
    3D ResNet Block & kernel $3 \times 3$, stride 1, padding 1 & $B \times 256 \times 8 \times 8 \times 8$ & $B \times 256 \times 8 \times 8 \times 8$ \\
    \midrule
    Conv3D & kernel $3 \times 3$, stride 1, padding 1 & $B \times 256 \times 8 \times 8 \times 8$ & $B \times 512 \times 8 \times 8 \times 8$ \\
    \bottomrule
\end{tabular}
\vspace{\tabmargin}
\tablelabel{tab_lang_arch}
\end{table*}

\begin{table*}[t]
\caption{
    \textbf{Architecture for the ResNet2TSDF}.
}
\vspace{-2mm}
\centering
\setlength{\tabcolsep}{1.0em}
\small
\begin{tabular}{c ccc ccc} 
    \toprule
    Layer name & Weights/Parameters & Input size & Output size \\
    \midrule
    ResNet-18~\cite{he2016deep} &  - & $B \times 3 \times 256 \times 256$ & $B \times 512 \times 8 \times 8$ \\
    \midrule
    Linear\_to\_3D &  - & $B \times 512 \times 64$ & $B \times 512 \times 512$ \\
    \midrule
    3D ResNet Block & kernel $3 \times 3$, stride 1, padding 1 & $B \times 512 \times 8 \times 8 \times 8$ & $B \times 256 \times 8 \times 8 \times 8$ \\
    \midrule
    Upsample & kernel $3 \times 3$, stride 2, padding 0 & $B \times 256 \times 8 \times 8 \times 8$ & $B \times 256 \times 16 \times 16 \times 16$ \\
    \midrule
    3D ResNet Block & kernel $3 \times 3$, stride 1, padding 1 & $B \times 256 \times 16 \times 16 \times 16$ & $B \times 128 \times 16 \times 16 \times 16$ \\
    \midrule
    Upsample & kernel $3 \times 3$, stride 2, padding 0 & $B \times 128 \times 16 \times 16 \times 16$ & $B \times 128 \times 32 \times 32 \times 32$ \\
    \midrule
    3D ResNet Block & kernel $3 \times 3$, stride 1, padding 1 & $B \times 128 \times 32 \times 32 \times 32$ & $B \times 64 \times 32 \times 32 \times 32$ \\
    \midrule
    Upsample & kernel $3 \times 3$, stride 2, padding 0 & $B \times 64 \times 32 \times 32 \times 32$ & $B \times 64 \times 64 \times 64 \times 64$ \\
    \midrule
    3D ResNet Block & kernel $3 \times 3$, stride 1, padding 1 & $B \times 64 \times 64 \times 64 \times 64$ & $B \times 32 \times 64 \times 64 \times 64$ \\
    \midrule
    GroupNorm & num\_groups=32 & $B \times 32 \times 64 \times 64 \times 64$ &  $B \times 32 \times 64 \times 64 \times 64$  \\
    \midrule
    Conv3D &  kernel $1 \times 1$, stride 1, padding 0 & $B \times 32 \times 64 \times 64 \times 64$ & $B \times 1 \times 64 \times 64 \times 64$ \\
    \bottomrule
\end{tabular}
\vspace{\tabmargin}
\tablelabel{tab_resnet2tsdf_arch}
\end{table*}
\begin{table*}[t]
\caption{
    \textbf{Architecture for the ResNet2Voxel}.
}
\vspace{-2mm}
\centering
\setlength{\tabcolsep}{1.0em}
\small
\begin{tabular}{c ccc ccc} 
    \toprule
    Layer name & Weights/Parameters & Input size & Output size \\
    \midrule
    ResNet-18~\cite{he2016deep} &  - & $B \times 3 \times 256 \times 256$ & $B \times 512 \times 8 \times 8$ \\
    \midrule
    Linear\_to\_3D &  - & $B \times 512 \times 64$ & $B \times 512 \times 512$ \\
    \midrule
    3D ResNet Block & kernel $3 \times 3$, stride 1, padding 1 & $B \times 512 \times 8 \times 8 \times 8$ & $B \times 256 \times 8 \times 8 \times 8$ \\
    \midrule
    Upsample & kernel $3 \times 3$, stride 2, padding 0 & $B \times 256 \times 8 \times 8 \times 8$ & $B \times 256 \times 16 \times 16 \times 16$ \\
    \midrule
    3D ResNet Block & kernel $3 \times 3$, stride 1, padding 1 & $B \times 256 \times 16 \times 16 \times 16$ & $B \times 128 \times 16 \times 16 \times 16$ \\
    \midrule
    Upsample & kernel $3 \times 3$, stride 2, padding 0 & $B \times 128 \times 16 \times 16 \times 16$ & $B \times 128 \times 32 \times 32 \times 32$ \\
    \midrule
    3D ResNet Block & kernel $3 \times 3$, stride 1, padding 1 & $B \times 128 \times 32 \times 32 \times 32$ & $B \times 64 \times 32 \times 32 \times 32$ \\
    \midrule
    3D ResNet Block & kernel $3 \times 3$, stride 1, padding 1 & $B \times 32 \times 32 \times 32 \times 32$ & $B \times 32 \times 32 \times 32 \times 32$ \\
    \midrule
    GroupNorm & num\_groups=32 & $B \times 32 \times 32 \times 32 \times 32$ &  $B \times 32 \times 32 \times 32 \times 32$  \\
    \midrule
    Conv3D &  kernel $1 \times 1$, stride 1, padding 0 & $B \times 32 \times 32 \times 32 \times 32$ & $B \times 1 \times 32 \times 32 \times 32$ \\
    \bottomrule
\end{tabular}
\vspace{\tabmargin}
\tablelabel{tab_resnet2voxel_arch}
\end{table*}

\begin{figure*}[h!]
    \centering
    \includegraphics[width=0.95\linewidth]{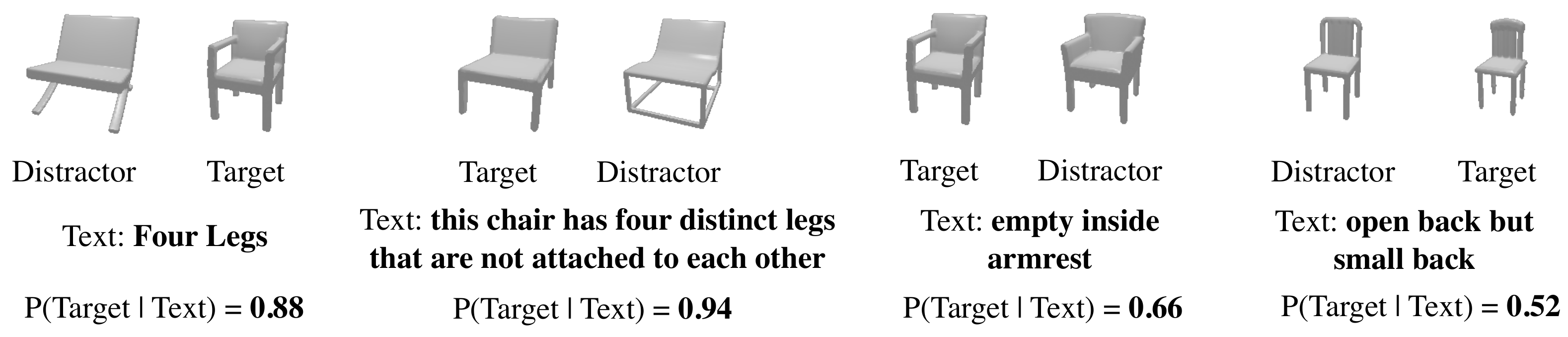}
    \vspace{-2mm}
    \caption{
    \textbf{Prediction from neural evaluator} Neural evaluator is trained to predict the target shape from two candidate shapes given a text description. Target indicates the true shape and P(Target | Text) indicates the prediction confidence for a given input. This highlights that neural evaluator can do a meaningful job in associating text descriptions to shapes. Low confidence in right most input instance indicates that the model confuses when input shapes are somewhat similar}
    \vspace \figmargin
    \figlabel{supp_lang_listener}
\end{figure*} 

\section{Implementation Detail}
We will release our code and learned models for reproducibility, but also describe here the experiments in additional detail.

\vspace{\subsecmargin}

\subsection{VQ-VAE Training}

\paragraph{Dataset Details.}

We train our proposed P-VQ-VAE using the objects from 13 categories of Shapenet~\cite{shapenet2015} data. These categories are [airplane, bench, cabinet, car, chair, display, lamp, speaker, rifle, sofa, table, phone, watercraft]. To extract SDF, we follow the preprocessing steps in DISN~\cite{Xu2019disn} and PixelTransformer~\cite{tulsiani2021pixel}. The shapes are normalized to lie in a origin-centered cube in $[-1,1]^3$, and their signed distance function is evaluated at locations in a uniformly sampled $64^3$ grid. We use $0.2$ as the threshold to further obtain the Truncated-SDF (T-SDF) representations.

\paragraph{Training Details.} Given a 3D shape $\bfX \in \mathbb{R}^{64^3}$, we first split $\bfX$ into $512$ patches of $\bfX_p \in \mathbb{R}^{8^3}$. After $E_{\psi}$ encodes each patch independently, we have the latent code $\hat{z}_{\bfi}$ for each patch at location $\bfi$. We perform vector quantize step $VQ$ for all $\hat{z}_{\bfi}$ to obtain $z_{\bfi}$.  Gathering $\hat{z}_{\bfi}$ and $z_{\bfi}$ for all location $\bfi$ into grids gives us the complete latent space for $\hat{\bfZ}$ and $\bfZ$ respectively. Finally, $D_{\psi}$ decodes $\bfZ$ to output the reconstruction $\bfX^{\prime}$. We then use the training objective proposed in VQ-VAE work~\cite{oord2017neural}:
\begin{equation}
  \begin{aligned} \eqlabel{vq_loss}
  \mathcal{L}_{\mathrm{VQ-VAE}} = -\log p(X | \mathbf{Z} ) &+ \norm{\mathrm{sg}[\hat{\mathbf{Z}}] - \mathbf{Z}}^2 \\
                     &+ \norm{ \hat{ \mathbf{Z}} - \mathrm{sg}[\mathbf{Z}] }^2.
  \end{aligned}
\end{equation}
where the first term is the reconstruction loss and $\mathrm{sg}[\cdot]$ denotes the stop gradients. The second and third term in \eqref{vq_loss} is the $VQ$ objective and the commitment loss respectively.

\paragraph{Architecture Details.} We describe the details for P-VQ-VAE's encoder $E_{\psi}$ at~\tableref{tab_enc}, decoder $D_{\psi}$ at~\tableref{tab_dec}. For the hyperparameters of the Codebook $\mathcal{Z}$, we use number of codebook entries 512, and the dimensionality of $z \in \mathbb{R}^{256}$.

\subsection{Non-sequential Autoregressive Modeling}

\paragraph{Training Details.} Given the learned P-VQ-VAE, we train our non-sequential autoregressive model over the low-dimensional shape space. Given a 3D shape $\bfX$, we first use $E_{\psi}$ and the codebook $\mathcal{Z}$ to obtain its discrete representation $\bfZ$ in the latent space.
Given this low-dimensional and discrete representation, we autoregressively model  $p(\bfZ)$ using a randomly permuted order of latent variables $\{z_{\mathbf{g}_1}, z_{\mathbf{g}_2}, z_{\mathbf{g}_3}, \cdots\}$ for factorizing the joint distribution:
\begin{equation}
    \begin{aligned}
        p(\mathbf{Z}) = p_{\theta}(z_{\mathbf{g}_1} | \varnothing) \cdot p_{\theta}(z_{\mathbf{g}_2} | z_{\mathbf{g}_1}) \cdot p_{\theta}(z_{\mathbf{g}_3} | z_{\mathbf{g}_1},z_{\mathbf{g}_2}) \cdots
    \end{aligned}
\end{equation}
In our implementation, the input to the transformer \textbf{T} are all the elements from $\{z_{\mathbf{g}_1}, z_{\mathbf{g}_2}, \cdots, z_{\mathbf{g}_{511}} \}$, and the query locations for the next elements $\{\bfi_{\mathbf{g}_2}, \bfi_{\mathbf{g}_3}, \cdots, \bfi_{\mathbf{g}_{512}} \}$. The outputs of \textbf{T} are the probability distributions over the codebook elements. For instance, to predict the $\mathbf{g}_i$-th element in the sequence, we have
\begin{equation}
    \begin{aligned}
        p(z_{\mathbf{g}_i}) = \textbf{T}(\{z_{\mathbf{g}_1}, z_{\mathbf{g}_2}, \cdots, z_{\mathbf{g}_{<i}} \}, \bfi_{\mathbf{g}_i})\, .
    \end{aligned}
\end{equation}
In practice, the training is done in parallel and the targets are the elements starting from $\mathbf{g}_2$:  $\{z_{\mathbf{g}_2}, z_{\mathbf{g}_3}, \cdots, z_{\mathbf{g}_{512}} \}$. We feed the attention mask with upper-triangular matrix of $- \infty$, and zeros on the diagonal to make sure the information do not leak from the future elements. We use fourier features for the positional embedding for all locations $\bfi$ following Tancik~\etal~\cite{tancik2020fourier}. The training objective is minimizing the expected negative log-likelihood, where we sample random orders or variables in every iteration:
\begin{equation}
    \begin{aligned}
        \mathcal{L}_{\mathrm{\textbf{T}}} = \mathbb{E}_{\bfX \sim p(\bfX)} - [ \log p(\bfZ) ]\, .
    \end{aligned}
\end{equation}

\paragraph{Inference.} During inference, the transformer can achieve autoregressive generation by completing either an empty sequence (unconditional generation) or incomplete sequence (shape completion). Given an arbitrary set $\mathbf{O} \equiv (\{\mathbf{g}_j\}_{j=1}^{k})$ of observed latent variables and their locations, we repeated apply
\begin{equation}
    \begin{aligned}
        p(z_{\mathbf{g}_{t}}) = \textbf{T}(\{\mathbf{O}, \hat{z}_{\mathbf{g}_{k+1}}, \cdots, \hat{z}_{\mathbf{g}_{t-1}}\}, \bfi_{\mathbf{g}_{t}})\, .
    \end{aligned}
\end{equation}
for $t = k+1, k+2, \cdots, 512$. $\hat{z}_{t}$ is obtained by sampling from the probability distribution $p(z_{\mathbf{g}_{t}})$. Once we have the complete sequence, we feed the estimated $\bfZ$ into $D_{\psi}$ to output the 3D shape.

\paragraph{Architecture Detail.} We adopt a transformer based architecture to learn the autoregressive prior over 3D shapes. Our model consists of 12 Encoder layers, with 12 multi-head attention heads and a hidden dimension of 768. In similar spirit as BERT~\cite{devlin2018bert} our transformer does not contain a `Decoder' i.e. all attention layers are self-attention.  We directly reuse the learned codebook for the input embedding of each token.

\subsection{Naive Condition Predictions}

\paragraph{Image Inference Module: Training Details.} Given a pair of image and its corresponding 3D shape $(I, \bfX)$, we first use a pretrained ResNet-18 to extract the image features. After that, we adopt a linear layer to lift the image features into 3D followed by several UpConv3D layers to output $8^3$ grids. Each grid contains the probability distribution $p(\mathbf{z}_{\mathbf{i}} | I)$ over the codebook entries learned by the P-VQ-VAE. We train the image inference module using a cross entropy loss, where the ground truth classes is given by $\bfZ = E_{\psi}(\bfX)$. The architecture details for image inference module is shown at~\tableref{tab_img_arch}.

\paragraph{Language Inference Module.} We encode text descriptions using a pre-trained BERT~\cite{devlin2018bert} encoder. Since text can be of varying length, we follow a common practice in language classification tasks and use the pooler output: a linear transformation over first token's representation as the output for learning the conditional. Following a similar architecture as image conditioning, we adopt linear layers to lift text-features into 3D. A group of UpConv3D layers are further used to produce the $8^3$ grid. This module is trained with cross entropy loss and exact architecture details are included in \tableref{tab_lang_arch}. 

\paragraph{Weight parameter $\alpha$ between the shape prior and the conditional marginal.} For the per-location condition, we use a hyperparameter $\alpha$ to balance the condition between the shape prior and the conditional marginals. Specifically, in the equation we use to approximate the conditional distribution,
\begin{equation*}
    \begin{aligned} \eqlabel{cond_gen_alpha}
        \prod_j p(z_{\mathbf{g}_j} | z_{\mathbf{g}_{<j}},C) \approx \prod_j p_{\theta}(z_{\mathbf{g}_j} | z_{\mathbf{g}_{<j}})^{1 - \alpha} \cdot \prod_j p_{\phi}(z_{\mathbf{g}_j} | C)^{\alpha}
    \end{aligned}
\end{equation*}
we control the dependence on the conditioning $C$ by raising power of the shape prior to $1-\alpha$ and the condition marginal to $\alpha$, where $\alpha \in [0, 1]$. We use $\alpha=0.75$ across all experiments on image conditioning. For language conditional generation, we find $\alpha=0.5$ to work better.

\subsection{Experimental Details}

\paragraph{Details of the Baselines for Single-view 3D Prediction.}

We describe the training and architecture details in this paragraph.
\begin{compactitem}
    \item \textit{ResNet2TSDF}. Given a pair of image and its corresponding 3D shape $(I, \bfX)$, we use the similar architecture as the image inference module to extract the image features and lift them to 3D. However, we use extra UpConv3D layers to output $64^3$ grids instead, where each grid predicts the TSDF values. We train ResNet2TSDF with the L1 loss. 
    \item \textit{ResNet2Voxel}. We adopt the similar architecture of ResNet2TSDF, but the resulting outputs are the voxels prediction with resolution of $32^3$. ResNet2Voxel is trained with binary cross entropy loss where the targets are the ground truth voxels. 
\end{compactitem}
The architectures details for ResNet2TSDF and ResNet2Voxel are shown at~\tableref{tab_resnet2tsdf_arch} and~~\tableref{tab_resnet2voxel_arch}.

\paragraph{Details on Neural Evaluator} We leverage a Neural Listener architecture proposed by Achlioptas \etal~\cite{achlioptas2019shapeglot} to perform discriminative quantitative comparisons. The Neural Listener uses 2D rendered images and Point-Clouds (PC) to represent the corresponding 3D shape in Shapenet~\cite{shapenet2015}. Following the proposed implementation by Achlioptas \etal~\cite{achlioptas2019shapeglot}, we (1) finetune a VGG network for 8-way classification task of Shapenet objects: The 8 classes are [airplane, car, chair, lamp, rifle, sofa, table, watercraft] and gives $\sim 95\%$ accuracy on held out data; (2) train a self-supervised Point-Cloud AutoEncoder~\cite{achlioptas2018learning} with 2048 points under Chamfer loss~\cite{achlioptas2018learning} on chairs from Shapenet~\cite{shapenet2015}; (3) train a 3 layer LSTM to process VGG-features and text description; (4) train a MLP which takes in the PC-AE latent representation (128D) and output of LSTM as input and performs classification. We mainly digress from the original implementation by performing a 2-way classification with one target and one distractor as opposed to the original setting of on target and two distactors. The Neural Listener has a classification accuracy of $\sim83\%$ on held out data. \figref{supp_lang_listener} contains the visualizations indicating meaningful performance from neural evaluator.
\paragraph{Joint Text-Shape Baseline} Our method proposes an approximate decomposition of joint probability of shapes and conditioning, into a product of an autoregressive prior over shapes and an independent conditional. In order to best highlight the efficacy of our proposed approach we create a baseline trained to directly learn this joint distribution. In further discussions, we call this baseline as JE. We imagine the task of text-to-shape generation as a Sequence to Sequence Translation task where a transformer Encoder (pre-trained using BERT~\cite{devlin2018bert}) is used to encode language. An autoregressive decoder is tasked with generating the latent representation which P-VQ-VAE can map back to 3D shapes. The Transformer Decoder uses the output of Encoder as \textit{memory} while making generations. For simplicity we assume a fixed rasterized order of generation (as opposed to the random order with which our proposed method is trained) and only use the chair category from Shapenet~\cite{shapenet2015} (i.e. the latent space from P-VQ-VAE was also only trained on Shapenet chairs). \figref{supp_lang_comp_base} shows that these assumptions help in generating higher-fidelity shapes however even then the proposed method's outputs align better with text query. 
\begin{figure*}[h!]
    \centering
    \includegraphics[width=0.95\linewidth]{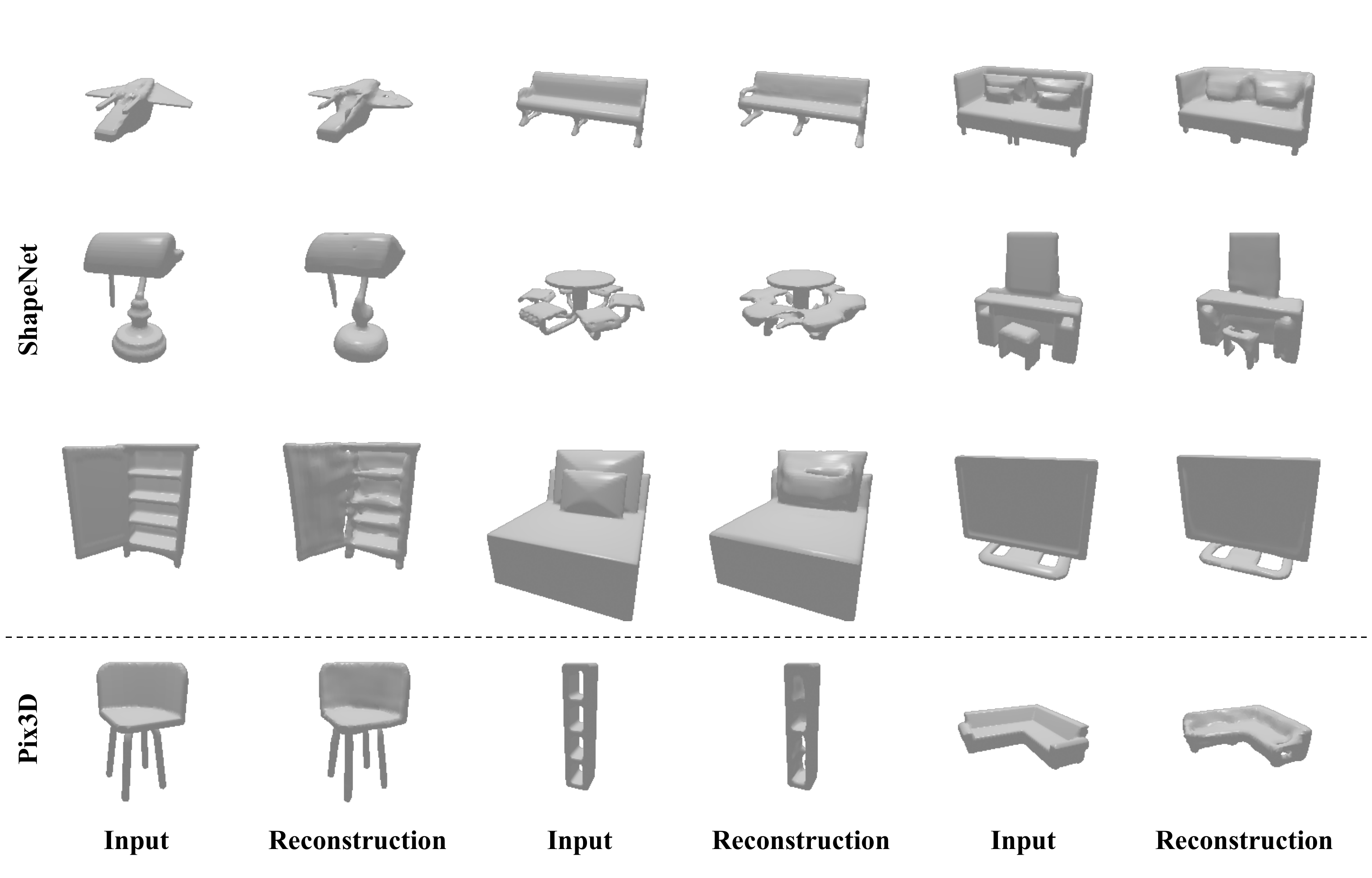}
    \vspace{-2mm}
    \caption{
    \textbf{Qualitative results for P-VQ-VAE's reconstruction.} We analyze the expressivity of our low-dimensional shape representation by visualizing the input shape (left) and the shape decoded from the encoded latent representation (right).}
    \vspace \figmargin
    \figlabel{supp_vqvae}
\end{figure*} 
\begin{figure*}[h!]
    \centering
    \includegraphics[width=0.8\linewidth]{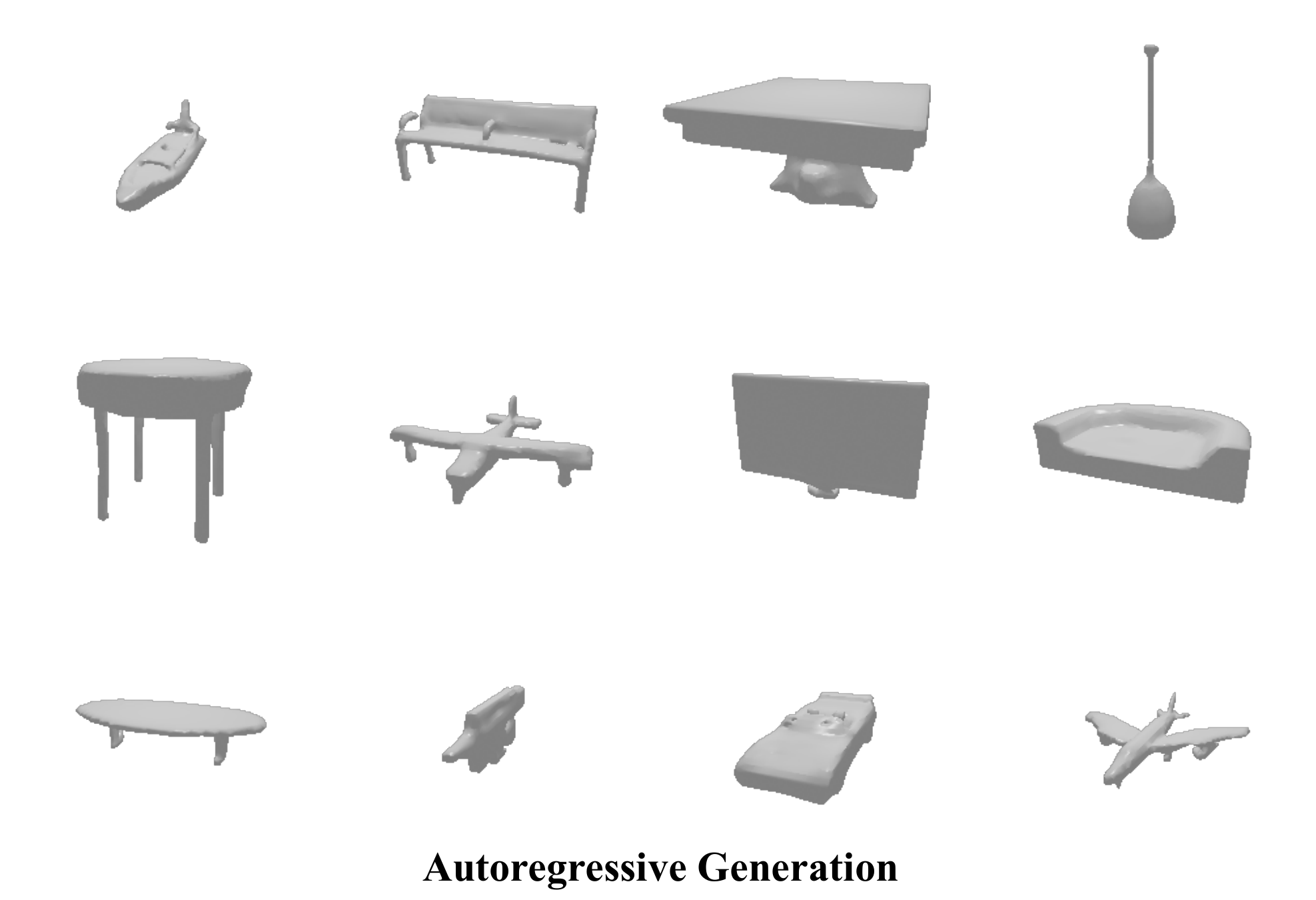}
    \vspace{-2mm}
    \caption{
    \textbf{Qualitative results of autoregressive generation from the transformer.} We validate that the transformer learns the representation of generic shapes. We achieve the generation by autoregressively predicting the next tokens starting from an empty sequence.}
    \vspace \figmargin
    \figlabel{supp_tf_gen}
\end{figure*} 

\section{Autoregressive Results}
\vspace{\subsecmargin}

\paragraph{Reconstruction of P-VQ-VAE.} In~\figref{supp_vqvae}, we demonstrate that P-VQ-VAE can faithfully reconstruct the given input shape $\bfX$. Although P-VQ-VAE is trained on the ShapeNet, it can reconstruct the input shape from Pix3D as well.

\paragraph{Random Generation of the Non-sequential Transformer.} We show the unconditional generation results with the transformer at~\figref{supp_tf_gen}. The generation is achieved by autoregressive generation starting from an empty sequence. The diverse results across all categories show that the non-sequential autoregressive prior learns the representation of the generic shapes.

\begin{figure*}[h!]
    \centering
    \includegraphics[width=0.95\linewidth]{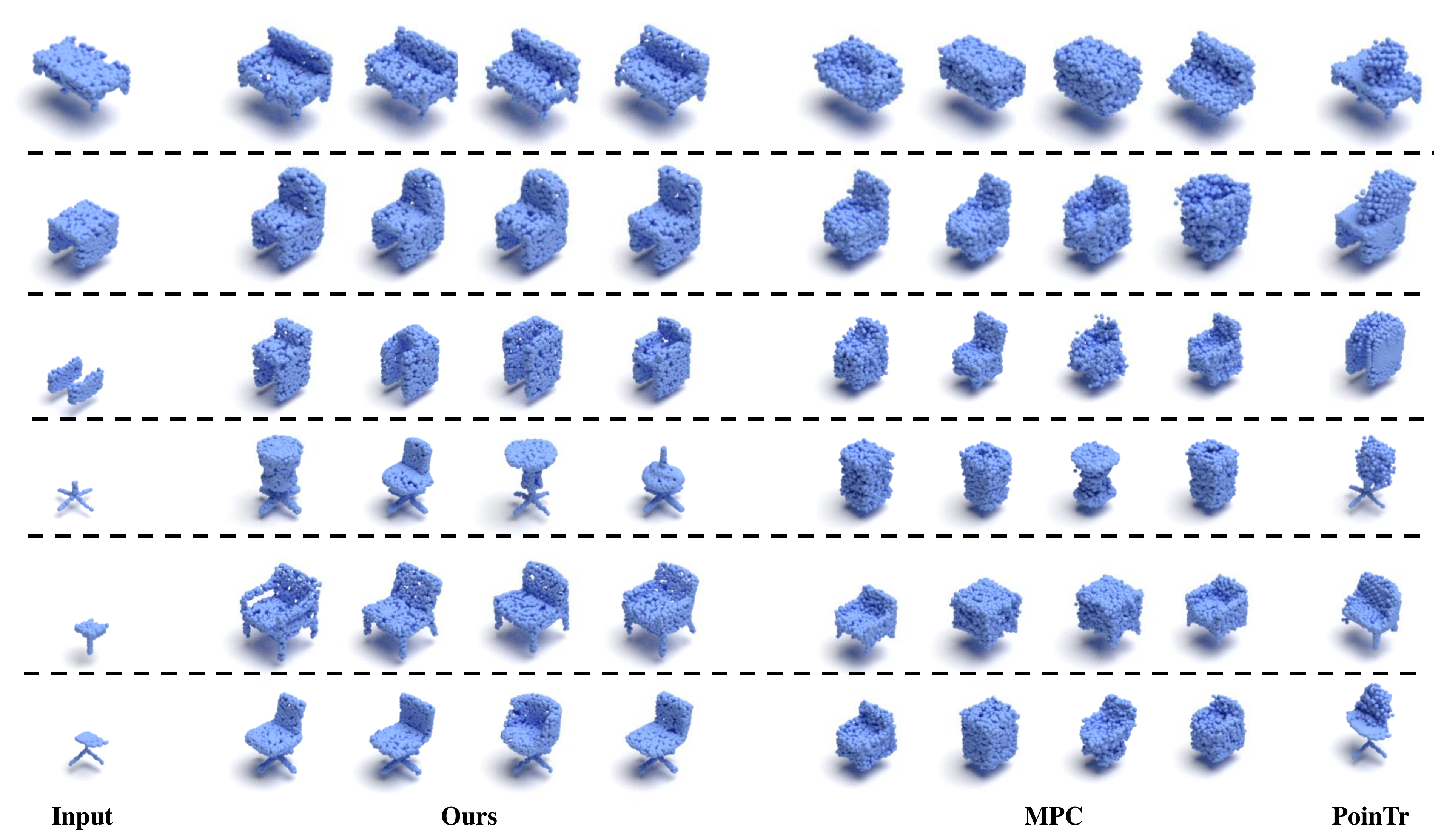}
    \vspace{-2mm}
    \caption{
    \textbf{More comparisons with competing methods for the shape completion.}
    }
    \vspace \figmargin
    \figlabel{supp_shape_comp_base}
\end{figure*} 
\begin{figure*}[h!]
    \centering
    \includegraphics[width=0.95\linewidth]{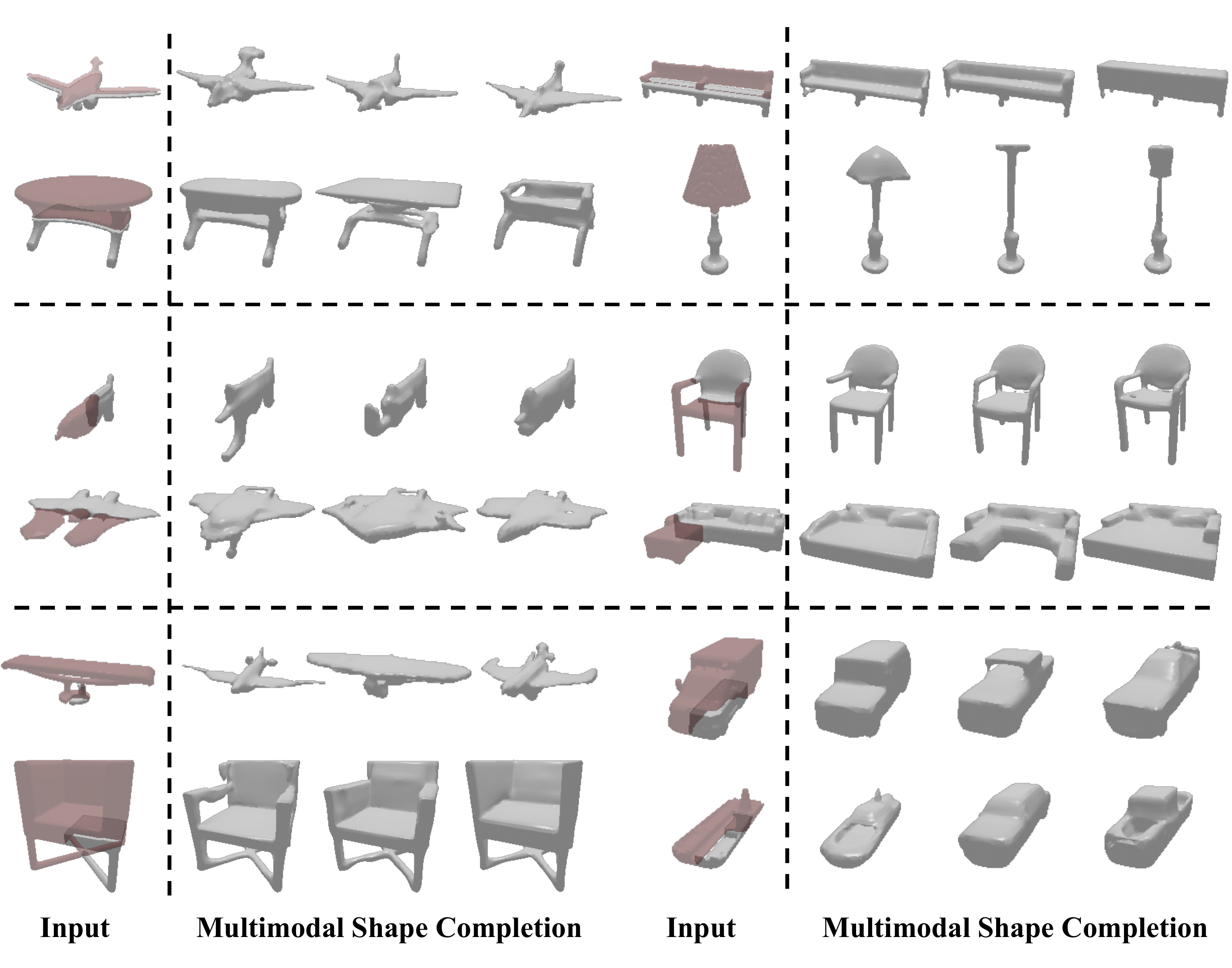}
    \vspace{-2mm}
    \caption{
    \textbf{More shape completions results with the proposed method.} Given the partial inputs, the proposed approach is able to generate diverse plausible 3D shapes consistent with the partial input. For example in row-5, a table like structure is reconstructed as an aeroplane. \red{Red cuboid} denotes the missing parts.}
    \vspace \figmargin
    \figlabel{supp_shape_comp_ours}
\end{figure*} 

\section{Shape Completion Results}
\vspace{\subsecmargin}

\paragraph{More Comparison of Multimodal Shape Completion.}
We show more comparisons of shape completion with baselines at \figref{supp_shape_comp_base}.

\paragraph{More Results of Multimodal Shape Completion.}
We show more results of shape completion at \figref{supp_shape_comp_ours}.
\begin{figure*}[h!]
    \centering
    \includegraphics[width=0.8\linewidth]{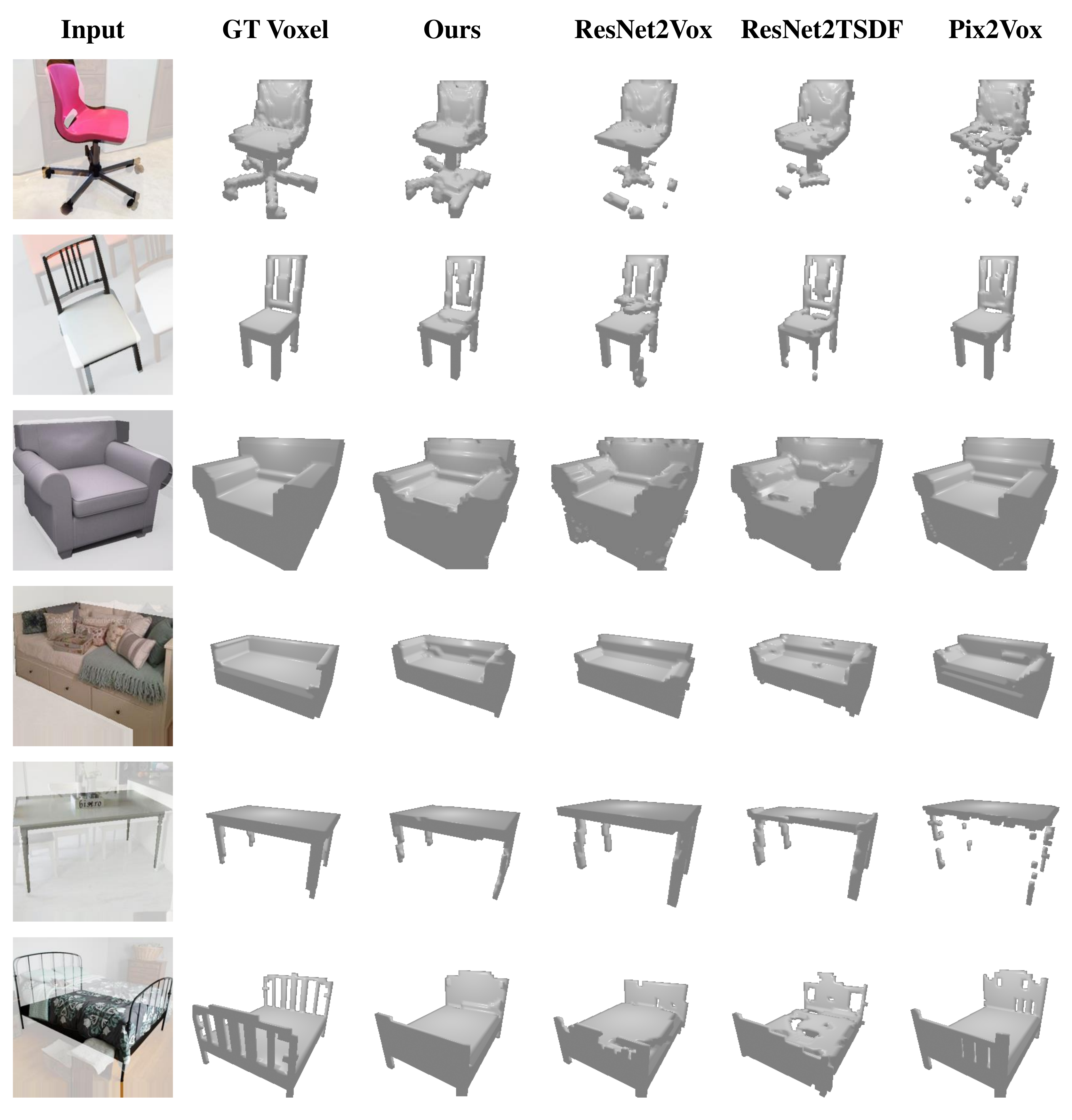}
    \vspace{-2mm}
    \caption{ \textbf{More comparisons with the competing methods on single-view reconstruction.} Given an image as input, we show the single-view reconstruction results with the proposed method and how it compares against other competing methods. }
    \vspace \figmargin
    \figlabel{supp_img2shape_comp}
\end{figure*} 
\begin{figure*}[h!]
    \centering
    \includegraphics[width=0.8\linewidth]{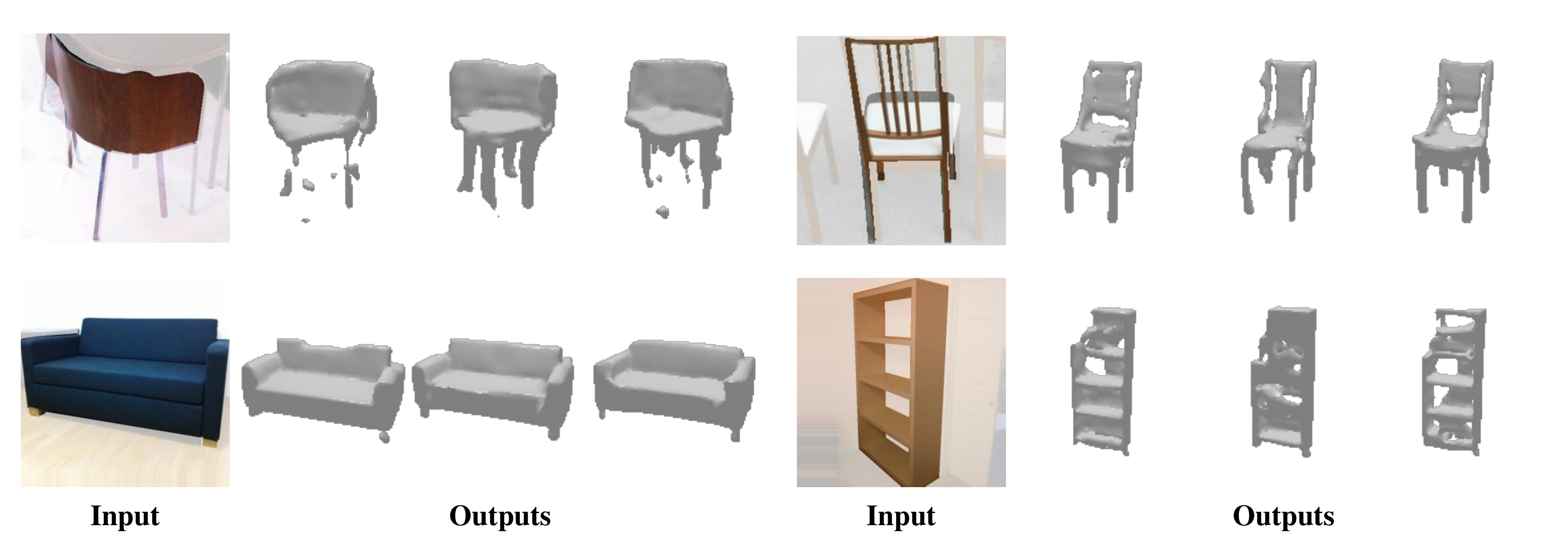}
    \vspace{-2mm}
    \caption{
    \textbf{More single-view reconstruction results from the proposed methods.} Given an image as input, we show the multimodal generation results with the proposed method.}
    \vspace \figmargin
    \figlabel{supp_img2shape_ours}
\end{figure*} 

\section{Single-view Results}
\vspace{\subsecmargin}
We show more single-view reconstruction results in this section. We provide more comparisons with the competing method at~\figref{supp_img2shape_comp}, and more results from the proposed method at~\figref{supp_shape_comp_ours}.

\section{Language-guided Generation Results}
\vspace{\subsecmargin}
\begin{figure*}[h!]
    \centering
    \includegraphics[width=0.95\linewidth]{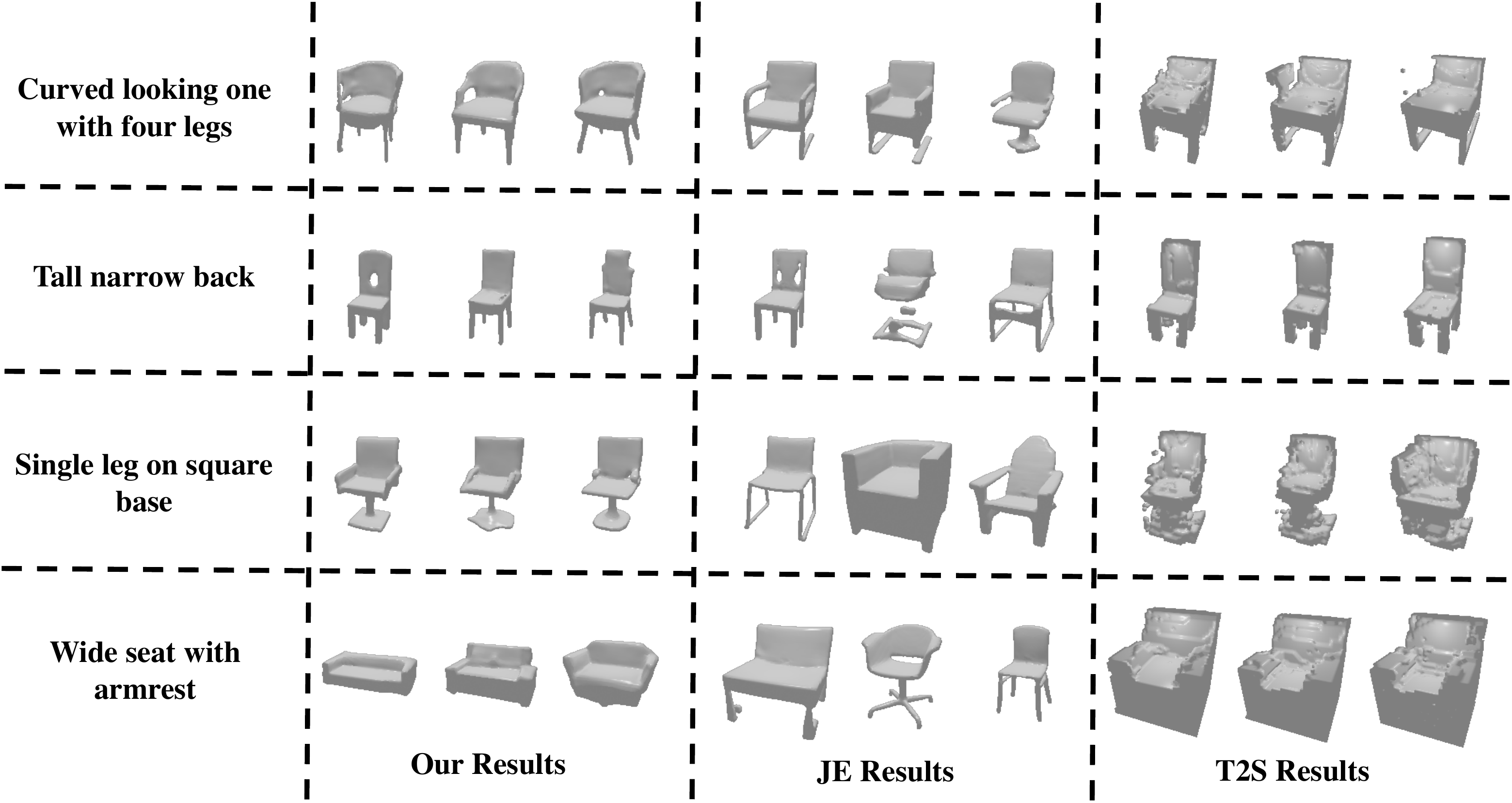}
    \vspace{-2mm}
    \caption{
    \textbf{Qualitative Comparison with baselines for language based generation.} While comparing our proposed method (Ours) with two baselines (JE and T2S) we report three random generations from each model for every text description. We observe that T2S shape generations are better aligned with text as compared to JE. Our proposed method is able to do both (a) align better with conditioning and (b) generate realistic 3D shapes. }
    \vspace \figmargin
    \figlabel{supp_lang_comp_base}
\end{figure*} 
\paragraph{Comparisons with Baseline}
\figref{supp_lang_comp_base} contains qualitative results for comparing our proposed approach with the two baselines (JE and Text2Shape (T2S)~\cite{chen2018text2shape}). For every text description, we include three random generations for each of the methods. The visualized images for ours and JE are rendered from $64^3$ T-SDFs while for T2S they are rendered from $32^3$ voxels. We observe that results from our approach are more aligned to corresponding text. We also observe the lack of diversity in the generations from T2S. We suspect this could be because of Mode Collapse while fine-tuning the GAN in T2S~\cite{chen2018text2shape} on data from ShapeGlot~\cite{achlioptas2019shapeglot}.
\begin{figure*}[h!]
    \centering
    \includegraphics[width=0.95\linewidth]{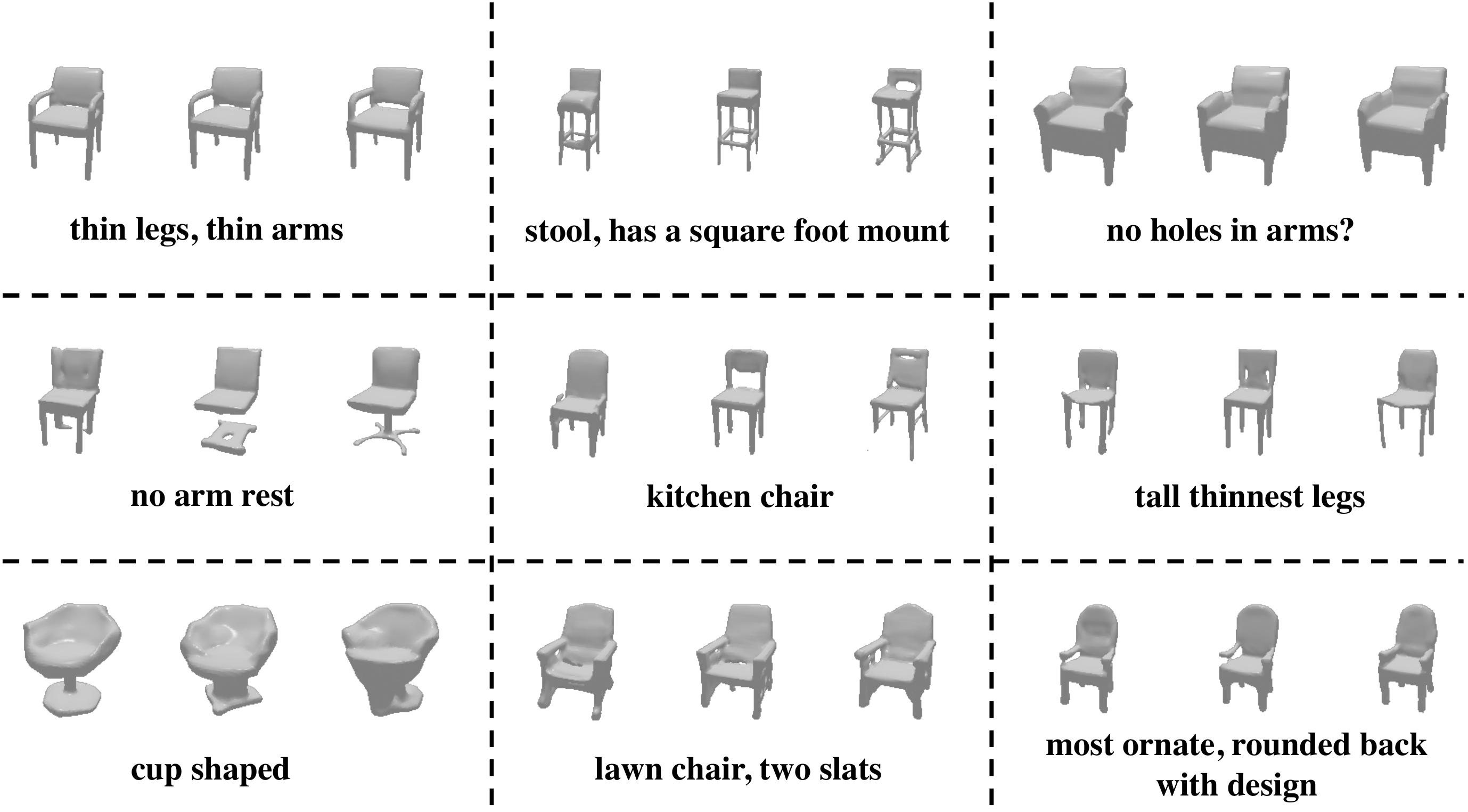}
    \vspace{-2mm}
    \caption{
    \textbf{Qualitative results of language based generations from our proposed approach} We report 3 random shapes generated for a given text description (in bold). We observe that while generated shapes are often meaningful, then can sometimes be random, example a flying chair in Row-2 Column-2. Bottom row includes some arguably rare text descriptions from held out data.}
    \vspace \figmargin
    \figlabel{supp_lang_more_res}
\end{figure*} 
\paragraph{Qualitative results} \figref{supp_lang_more_res} includes results of Text Conditioned Shape Generation on diverse text descriptions from held out data. For each text we include three random generations. The bottom row of \figref{supp_lang_more_res} includes some complex descriptions and highlight that generation quality is impacted with such complex and ambiguous descriptions.

\end{document}